\documentclass[review]{elsarticle}
\usepackage{hyperref}
\usepackage{amsmath,amssymb,amsfonts}
\usepackage{algorithmic}
\usepackage{algorithm}
\usepackage{textcomp}
\usepackage[T1]{fontenc}
\usepackage{float}
\usepackage{subfig}
\usepackage{amsmath}
\usepackage{geometry}
\usepackage{multirow}

\usepackage{booktabs}
\usepackage{stfloats}

\usepackage{color}
\usepackage[dvipsnames, svgnames, x11names]{xcolor}
\usepackage[capitalize]{cleveref}
\Crefname{section}{Section}{Sections}
\crefname{figure}{Fig.}{Figs.}
\Crefname{table}{Table}{Tables}
\journal{Journal of \LaTeX\ Templates}

\begin{document}

\begin{frontmatter}

\title{eX-ViT: A Novel eXplainable Vision Transformer for Weakly Supervised Semantic Segmentation}

\author[mymainaddress]{Lu Yu}
\ead{lu.yu@my.jcu.edu.au}

\author[mysecondaryaddress]{Wei Xiang\corref{mycorrespondingauthor}}
\cortext[mycorrespondingauthor]{Corresponding author}
\ead{w.xiang@latrobe.edu.au}

\author[mythirdaddress]{Juan Fang}
\ead{fangjuan@bjut.edu.cn}

\author[mysecondaryaddress]{Yi-Ping Phoebe Chen}
\ead{phoebe.chen@latrobe.edu.au}

\author[mysecondaryaddress]{Lianhua Chi}
\ead{L.Chi@latrobe.edu.au}

\address[mymainaddress]{College of Science and Engineering, James Cook University, Cairns, QLD 4878, Australia}
\address[mysecondaryaddress]{School of Computing, Engineering and Mathmatical Sciences, La Trobe University, Melbourne, VIC 3086, Australia}
\address[mythirdaddress]{College of Information Technology, Beijing University of Technology, Beijing 100124, China}

\begin{abstract}
Recently vision transformer models have become prominent models for a range of vision tasks. These models, however, are usually opaque with weak feature interpretability, making their predictions inaccessible to the users. While there has been a surge of interest in the development of post-hoc solutions that explain transformer decisions, these methods can not be broadly applied to different transformer architectures, as rules for interpretability have to change accordingly based on the heterogeneity of data and model structures. Moreover, there is no method currently built for an intrinsically interpretable transformer, which is able to explain its reasoning process and provide a faithful explanation. To close these crucial gaps, we propose a novel vision transformer dubbed the eXplainable Vision Transformer (eX-ViT), an intrinsically interpretable transformer model that is able to jointly discover robust interpretable features and perform the prediction. Specifically, eX-ViT is composed of the Explainable Multi-Head Attention (E-MHA) module, the Attribute-guided Explainer (AttE) module and the self-supervised attribute-guided loss. The E-MHA tailors explainable attention weights that are able to learn semantically interpretable representations from local patches in terms of model decisions with noise robustness. Meanwhile, AttE is proposed to encode discriminative attribute features for the target object through diverse attribute discovery, which constitutes faithful evidence for the model’s predictions. In addition, a self-supervised attribute-guided loss is developed for our eX-ViT, which aims at learning enhanced representations through the attribute discriminability mechanism and attribute diversity mechanism, to localize diverse and discriminative attributes and generate more robust explanations. As a result, we can uncover faithful and robust interpretations with diverse attributes through the proposed eX-ViT. To verify and evaluate our method, we apply the eX-ViT to several weakly supervised semantic segmentation (WSSS) tasks, since these tasks typically rely on accurate visual explanations to extract object localization maps. Particularly, the obtained explanation results obtained via eX-ViT are regarded as pseudo segmentation labels to train WSSS models. Comprehensive simulation results demonstrate that the proposed eX-ViT shows comparable performance to the supervised baselines, and outperforms the state-of-the-art black-box methods using only image-level labels in accuracy and interpretability. 
\end{abstract}


\end{frontmatter}

%\linenumbers

\section{Introduction}
\label{sec:intro}
Over the last few years, transformer models have attracted increasing attention with encouraging results in a number of challenging domains, such as natural language processing, vision or graphs \cite{xumulti2022}. The Multi-Head Attention (MHA) and Multi-Layer Perceptron (MLP) modules in transformers effectively model global representations without convolution \cite{dosovitskiy2020ViT}. This framework is proved to be effective to capture long-range dependencies for image recognition. Despite their excellent performance, most transformer architectures are usually expressed as black boxes \cite{ruaffinity2022}. To be concrete, transformers can achieve impressive results but lack an explanation of the results due to the large number of parameters and the complex interactions between modules. Since transformers have high applicability to high-risk decision-making domains such as healthcare and autonomous driving, there is a strong necessity for gaining insights in model's decisions \cite{Chen02909}. Besides, the great popularity of transformer models necessitates the explainable solutions to their inference process. An interpretable solution is able to aid in debugging the models and identifying crucial modules for downstream tasks \cite{Hao0W021}.

Explainable Artificial Intelligence (XAI) is an emerging sub-field of AI pursuing to capture the properties that have influence over the decision of a model \cite{arrieta2020explainable}. Depending on the phases where predictions and explanations are performed, these methods can be categorized into two types: intrinsically explainable models and post-hoc explanation methods. Unlike post-hoc models, explainable models aim to directly incorporate interpretability in the structure of the models, thus revealing the intrinsic reasoning process of the models. Several previous studies have pointed out that explainable models outperform post-hoc methods in faithfulness and stability \cite{WangW21}. Unfortunately, little work has been done so far in the field of explainable transformers. Pan et al. \cite{PanPJWFO21} developed an interpretability-aware method to drop less informative patches for transformers to reduce computational cost. Caron et al. \cite{Caron2021} proposed a self-supervised approach based on vision transformers and concluded that the attentions maps contain features about the semantic information of the image. However, the explicit expressive features were not explored to obtain faithful explanations w.r.t. model decisions. 

In contrast to intrinsically explainable models, post-hoc explanation methods aim to obtain explanations by approximating a trained model and its predictions. For instance, Zhou et al. \cite{zhou2016learning} introduced a technique called Class Activation Mapping (CAM) for identifying informative areas w.r.t. the model decisions by a certain type of CNNs which do not have fully-connected layers. Various studies have been made on post-hoc frameworks to uncover the predictions of the transformers. Brunner et al. \cite{BrunnerLPRCW20} recently developed a method to visualize the extraction of the attention heads. Abnar et al. \cite{AbnarZ20} attempted to build elaborate explanations across layers of the transformer. Hao et al. \cite{Hao0W021} developed a self-attention attribution mechanism to explain the information flow inside transformers. However, it is known to be unfaithful and unreliable since irrelevant tokens or less important attention heads often become highlighted. Another option is to employ the layer-wise relevance propagation (LRP) method to interpret transformer models. Chefer et al. \cite{CheferLRP21} proposed a layer-wise relevance propagation (LRP) method to compute different relevance of the attention heads throughout the transformer. Voita et al. \cite{VoitaTMST19} used LRP to calculate relevance for each attention head in each transformer layer. However, the LRP mechanism was done in a limited way, in which different relevance rules are subject to the heterogeneity of different data and model structures \cite{WangW21}. Moreover, post-hoc methods cannot faithfully uncover the decision making process of the trained models, and the explanations can be easily impaired by different input schemes (e.g., perturbations or transformations). Therefore, post-hoc methods are considered to be more fragile, sensitive, and less faithful.

Against the above background, this paper aims to design the so-called eXplainable Vision Transformer (eX-ViT) with the inherent attribute of explainability and high performance. Specifically, the eX-ViT comprises the Explainable Multi-Head Attention (E-MHA) module, which can inherently provide interpretable attention maps that align with informative input patterns with noise robustness. Furthermore, the Attribute-guided Explainer (AttE) module is integrated into the eX-ViT, to learn discriminative attribute features for the target object. Intuitively, we assume each object is made up of several attributes, which could be basic elements including color, shape, and texture, or higher-level local features such as body parts. Our key idea is to decompose the feature representation into a set of learnable attribute features for the target object, capable of capturing diverse and discriminative object features. Besides, a novel attribute-guided loss is designed to promote the learning process inside AttE in a self-supervised manner. More precisely, this loss implicitly adds the regularization to force the representations to focus on various attributes of each target class through the attribute discriminability mechanism and attribute diversity mechanism. We then verify and evaluate our method on several weakly supervised semantic segmentation (WSSS) tasks, as they typically rely on accurate visual explanations to extract object localization maps. Particularly, the obtained interpretation results obtained via eX-ViT are regarded as pseudo segmentation labels to train WSSS models. To the best of our knowledge, this is the first work to develop an intrinsically explainable vision transformer for WSSS tasks. In summary, the major contributions of this paper are:
\begin{itemize}
    \item We propose a novel eXplainable Vision Transformer (eX-ViT), which provides faithful and robust explanations with model-inherent interpretability. Specifically, the proposed eX-ViT is able to provide explainable representations with comparable or better performance than state-of-the-art transformers (e.g., Swin Transformer \cite{liu2021swin} and PVT \cite{wang2021pvt});
    \item We propose a novel Explainable Multi-Head Attention (E-MHA) module, which, as a basic building block of the eX-ViT, has two key attributes. That is, it provides model-inherent explainable attention maps that align with the informative input patterns and is robust to noise;
    \item We propose the Attribute-guided Explainer (AttE) module, which is integrated into the eX-ViT to recognize diverse and discriminative object attribute features for the target object with only image-level labels through diverse attribute discovery;
    \item We propose the attribute-guided loss function, which enables the self-supervised learning in the proposed eX-ViT, capable of not only learning explanations that are faithful to the model's predictions, but also resulting in more robust feature representations across data transformations;
    \item Comprehensive simulation results demonstrate that the proposed eX-ViT is comparable to the supervised baselines, and outperforms the state-of-the-art transformers in accuracy and interpretability.
\end{itemize}

The remainder of the paper is organized as follows. \cref{sec2} briefly describes some recent related works on vision transformers, XAI techiniques for transformers, and weakly supervised semantic segmentation methodologies. \cref{sec3} presents the explainable architecture, i.e., eX-ViT, and introduces its main modules. Experimental results and discussions are presented in \cref{sec:experi}, followed by concluding remarks drawn in \cref{sec5}.

\section{Related work}
\label{sec2}
\subsection{Transformers for vision}
Transformer-based models have recently been introduced to vision tasks and achieved remarkable progress. One of purely transformer-based models is the ViT \cite{dosovitskiy2020ViT}, which has exhibited impressive performance without convolution. However, the ViT is inferior to CNNs when trained on a mid-sized dataset. DeiT \cite{DeiT2021} and T2T-ViT \cite{yuan2021tokenstotoken} addressed this issue by designing efficient transformer-based backbones. Liu et al. \cite{liu2021swin} proposed the Swin Transformer to compute the local attention within a window for object detection tasks. Wang et al. \cite{wang2021pvt} designed the PVT, which is another architecture that has a feature hierarchy with multiple stages and multiple resolutions for various vision tasks. Moreover, there are some research studies with modified ViT architectures that benefit downstream vision tasks such as semantic segmentation. However, most of existing designs focus on efficient and effective frameworks for downstream tasks without considering interpretability. Thus these methods tend to be less faithful to the users. Recently, Wei et al. \cite{gaoTSCAM2021} proposed a transformer-based framework that generates semantic-aware CAMs for the weakly supervised object localization task. However, TS-CAM only leverages the class-agnostic feature maps and ignores the intrinsic semantic attributes for the target class. In this paper, we aim to address these issues by proposing the so-called eX-ViT, which exploits explainable features that are robust to noise and provides faithful explanations. 

\subsection{Explainable artificial intelligence (XAI) for transformers}
There are mainly two sub-fields of explainable techniques: intrinsically explainable models and post-hoc explanation methods. Unlike post-hoc models, intrinsically explainable models aim to directly incorporate interpretability in the structure of the models, thus revealing the intrinsic reasoning process of the models. Several previous studies have pointed out that explainable models outperform post-hoc methods in faithfulness and stability \cite{WangW21}. Unfortunately, little work has been done so far in the field of explainable transformers. Pan et al. \cite{PanPJWFO21} proposed an interpretability-aware method to drop less informative patches for transformers to reduce computational cost. However, the obtained interpretations were not complete due to the information loss. Recently, Caron et al. \cite{Caron2021} utilized a self-supervised approach called DINO based on vision transformers and concluded that the attentions maps contain features about the semantic information of the image. But the expressive features were not explored to obtain faithful explanations. Different from the majority of previous studies, we attempt to build the first explainable transformer architecture with the objective of learning interpretable features.

In terms of post-hoc explanation approaches, there are a variety of recent studies that explore the explainability for transformers. Chefer et al. \cite{CheferLRP21} proposed a layer-wise relevance propagation (LRP) method by introducing a relevancy propagation rule that is applicable to both positive and negative contributions. This approach, however, is not able to provide the interpretation for attention modules besides self-attention. Voita et al. \cite{voita2019analyzing} found that only a small portion of heads in transformers have a salient impact on the model decisions. Jawahar et al. \cite{JawaharSS19} proved that syntactic patterns are learned in lower blocks, while semantic patterns are gained through higher layers. Abnar et al. \cite{AbnarZ20} proposed to combine the attention scores across multiple layers, but this method fails to distinguish between positive and negative attributions. Raghu et al. \cite{RaghuUKZD21} analyzed the internal representations of vision transformers, and found out that they tend to learn more uniform representations across all layers. Recently, Chefer et al. \cite{CheferGW21} also proposed a generic approach to explain transformers including bi-modal ones. However, most of the existing post-hoc methods tend to be fragile, sensitive, and less faithful. Since they cannot faithfully uncover the decision making process of the trained models, and the explanations can be easily impaired by different input schemes (e.g., perturbations or transformations).

\subsection{Weakly supervised semantic segmentation}
Compared to supervised learning methods, WSSS aims at training models with weak labels such as bounding boxes and image-level labels. As the cornerstone of WSSS, most of existing designs in terms of WSSS tasks use the Class Activation Mapping (CAM) to extract object localization maps and approximate the segmentation mask. Despite the encouraging results, CAM suffers from the issue with incomplete object activation \cite{xumulti2022}. To address this drawback, several approaches have been proposed as the CAM expansions to remove the most discriminative parts of CAM and discover more complete object localization maps. Ahn et al. \cite{AhnK18} utilized the AffinityNet model to learn the semantic affinity between pixels, then propagated the CAM activations several times to revise CAMs. Yao et al. \cite{Yao0XZS0T021} accumulated the activated regions to discover the objects in the non-salient regions. Li et al. \cite{LiZLZZ21} proposed a group-wise learning task to discover similar semantic regions, forming more reliable pseudo ground-truth masks. Papandreou et al. \cite{PapandreouCMY15} adopted an Expectation-Maximization architecture to learn pseudo segmentation masks. Peng et al. \cite{PengKDAPD20} provided a method to train a CNN with discrete constraints and regularization priors for WSSS learning with favorable performance. The above methods are commonly based on CNNs, which reveals the inherent drawback of convolution. Xu et al. \cite{xumulti2022} introduced transformer attentions to learn class-specific localization maps. Ru et al. \cite{ruaffinity2022} adopted the semantic affinity in self-attentions in transformers to produce more integral pseudo labels for WSSS. However, there is still a large gap between fully supervised semantic segmentation and previous transformer-based WSSS methods. In our work, we propose a transformer-based model to extract explanation features to localize class-specific feature maps. We attempt to build a novel transformer architecture with the objective of learning interpretable representations in a self-supervised manner to narrow the supervision gap.

\begin{figure*}[t]
  \centering
      \includegraphics[width=0.9\textwidth, height=0.45\textwidth]{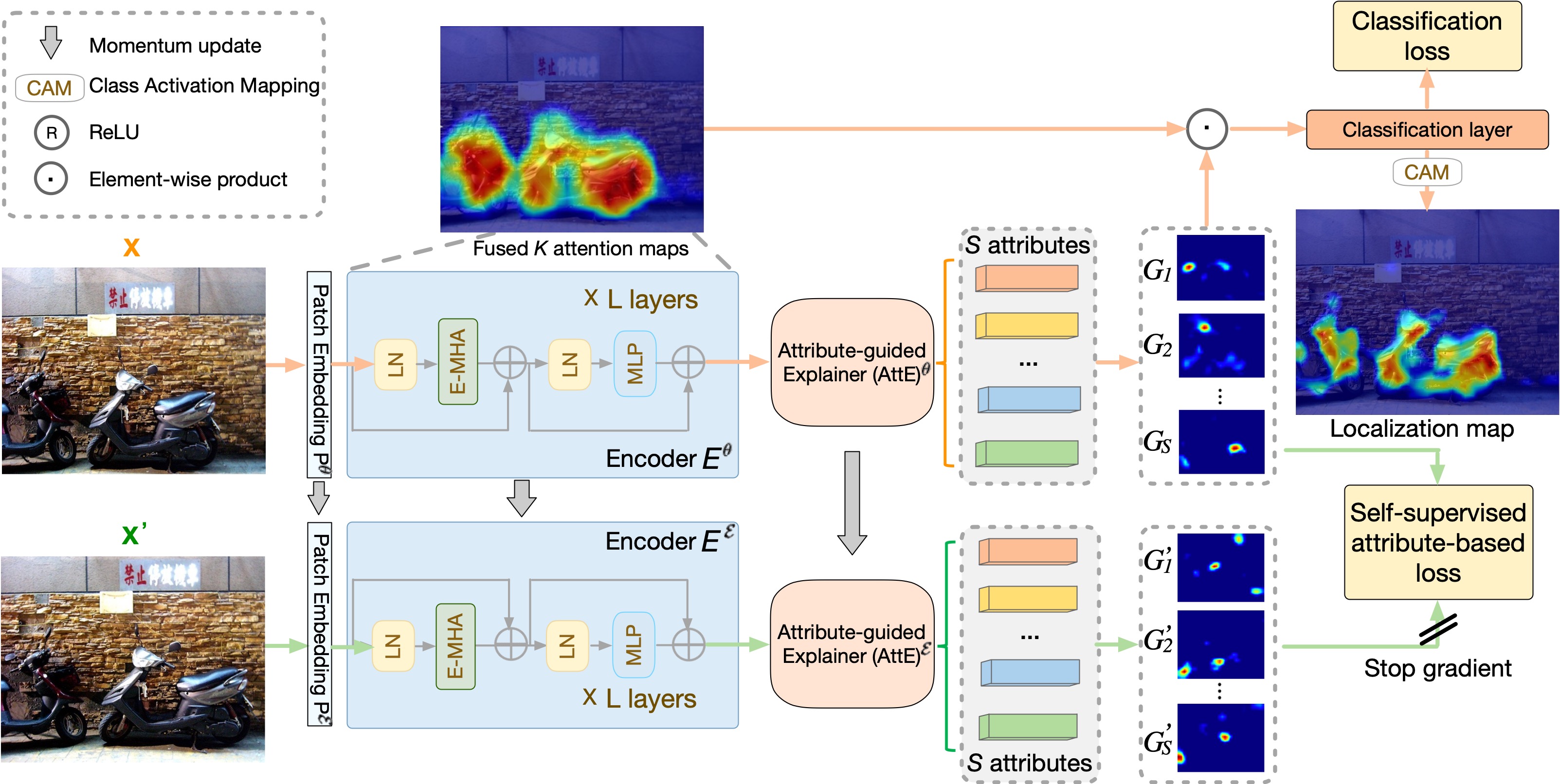}
  \caption{Illustration of the proposed eXplainable Vision Transformer (eX-ViT) architecture. $\boldsymbol{x}$ and $\boldsymbol{x}^{'}$ are two different random transformations of an input image. We use a transformer backbone as the encoder to extract feature maps, the backbone contains consecutive $L$ encoding layers with Explainable Multi-Head Attention (E-MHA) as the attention block. $\theta$ is the trainable module, while ${\mathcal{E}}$ is an exponential moving average of $\theta$. The Attribute-guided Explainer (AttE) is proposed atop of the encoder to decompose the attention maps into features of attributes through diverse attribute discovery, so as to facilitate the generation of more faithful and robust interpretations. We also design a self-supervised attribute-guided loss function for our eX-ViT, which aims at learning robust semantic representations via the attribute diversity mechanism and attribute discriminability mechanism.}
  \label{fig:fig1}
\end{figure*}

\section{Method}
\label{sec3}
This section details our proposed network architecture, i.e., the eX-ViT. First, we introduce the overall architecture. Second, we describe the intuition and design of the E-MHA and discuss several important properties of the E-MHA. Furthermore, The AttE is proposed to integrate into our eX-ViT to decompose the attention maps into features of attributes through diverse attribute discovery, and a self-supervised attribute-guided loss is adopted to learn robust semantic representations via the attribute diversity mechanism and attribute discriminability mechanism, which constitutes faithful evidence for the model’s predictions. 

\subsection{Architecture of the eX-ViT}
The overall architecture of our proposed eX-ViT is depicted in \cref{fig:fig1}. In particular, the eX-ViT is a siamese network, which comprises two branches for a pair of input images (two data augmentations from an original input) to learn interpretable attention maps in a self-supervised manner. Each branch comprises a transformer encoder with $L$ transformer layers consisting of the novel Explainable Multi-Head Attention (E-MHA) module, and the Attribute-guided Explainer (AttE) module atop of it. Following MoCo \cite{he2020momentum}, the parameters ${\mathcal{E}}$ in the lower branch uses a momentum update with the upper ${\theta}$. Empirically, the proposed architecture can conveniently replace the backbone networks in existing methods for WSSS tasks.

\subsection{Explainable Multi-Head Attention (E-MHA)}
\begin{figure}[t]
  \centering
  \includegraphics[width=1\linewidth, height=0.46\linewidth]{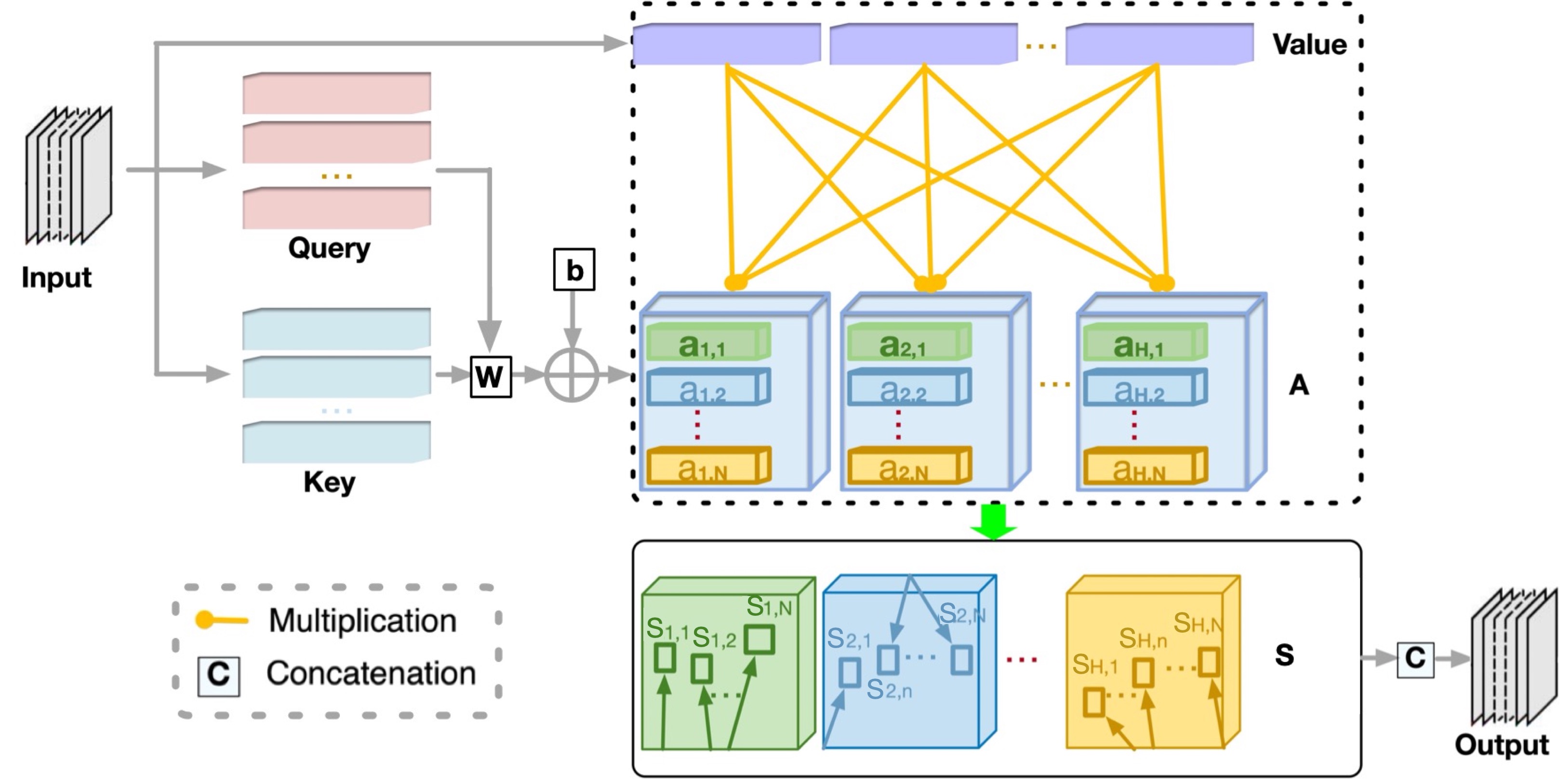}
  \caption{The architecture of Explainable Multi-Head Attention (E-MHA). Compared to MHA, the E-MHA dynamically aligns attention weights with the most discriminative input patterns during optimization, so as to be explainable and robust to noise.}
  \label{fig:fig2}
\end{figure}

In this section, we introduce our novel Explainable Multi-Head Attention (E-MHA) module as shown in \cref{fig:fig2}, which consists of $H$ parallel heads. Specifically, given an input feature map $\boldsymbol{X} \in \mathbb{R}^{T \times d}$ where $T$ is the spatial size and $d$ is the feature dimension, each head $H_h$ holds an explainable attention weight $\boldsymbol{A}_h \in \mathbb{R}^{N \times d}$ ($N$ is the spatial size of $\boldsymbol{A}_h$.) that represents relative importance of input features. That is, $\boldsymbol{A}_h$ aims to learn explainable features for the output through the proposed E-MHA module.

In particular, we structure this section around two crucial attributes of the E-MHA module:

{\bf Noise robustness:} The E-MHA is computed as a dynamic alignment between the input tokens and the attention weight. When the module is optimized, the attention weight is driven to focus on the most discriminative and class-related patterns from the input tokens. Instead of directly removing the irrelevant noises from the input image, we adopt a dynamic alignment mechanism in E-MHA to extract discriminative features from the input, thus reducing the noise information gradually and then preserving the key input patterns. This favorable attribute is empirically verified in section \cref{sec4.3}.

{\bf Inherent explainability:} Given input $\boldsymbol{X}$, the E-MHA aims to learn the attention weight that maximises the alignment between input tokens and the attention weight. During the optimization process, maximising this alignment means encoding the projected input values as eigenvectors of the attention weight. As a result of this property, the model-inherent attention weight is learned with the discriminative input patterns and thus directly used to explain model decisions without needing any external tools.

First, given input $\boldsymbol{X}$, the projected key, query and value are obtained as follows
\begin{equation}
    \boldsymbol{Q}=\boldsymbol{X} \boldsymbol{W}^Q, \quad \boldsymbol{K}=\boldsymbol{X} \boldsymbol{W}^K, \quad \boldsymbol{V}=\boldsymbol{X} \boldsymbol{W}^V,
    \label{eq1}
\end{equation}
where $\boldsymbol{W}^Q \in \mathbb{R}^{d \times d}$, $\boldsymbol{W}^K \in \mathbb{R}^{d \times d}$, and $\boldsymbol{W}^V \in \mathbb{R}^{d \times d}$ are trainable transform matrices. 

Second, the self-attention operation is constructed by
\begin{equation}
    \boldsymbol{W} = \frac{\boldsymbol{Q} \boldsymbol{K}^T}{\sqrt{d}},
    \label{eq2}
\end{equation}
where the obtained matrix $\boldsymbol{W}$ implies how much attention is paid on each token.

Third, the attention weight $\boldsymbol{A}$ is defined as follows
\begin{equation}
    \boldsymbol{A} = \ell(\boldsymbol{W}+\boldsymbol{b})^{\rm T},
    \label{eq3}
\end{equation}
where $\boldsymbol{b}$ is a trainable bias term, which is introduced as an initial alignment for the input patterns. $\ell(\cdot)$ is a non-linear function that scales the L2 norm of its input, i.e., $\ell(\boldsymbol{x})= \frac{\boldsymbol{x}}{||\boldsymbol{x}||_2}$ and $||\ell(\boldsymbol{x})|| \leq 1$.

In what follows, the self-attention feature $\boldsymbol{S}$ is formally expressed as 
\begin{equation}
    \boldsymbol{S} = \boldsymbol{A}^{\rm T}\boldsymbol{V},
    \label{eq4}
\end{equation}
According to \cref{eq3}, $||\boldsymbol{A}|| \leq 1$. Therefore $\boldsymbol{S}$ in \cref{eq4} is upper-bounded as follows
\begin{equation}
  \boldsymbol{S} = ||\boldsymbol{A}|| ~ ||\boldsymbol{V}|| \cos(\boldsymbol{A}, \boldsymbol{V}) \leq ||\boldsymbol{V}||.
  \label{eq5}
\end{equation}
When \cref{eq5} is optimized, the attention weight $\boldsymbol{A}$ is proportional to $\boldsymbol{V}$. In order to achieve maximal output, $\boldsymbol{A}$ is driven to align with the discriminative features in $\boldsymbol{V}$, instead of the uninformative noise. Therefore, $\boldsymbol{S}$ can only achieve this upper bound if all possible solutions of $\boldsymbol{v} \in \boldsymbol{V}$ are encoded as eigenvectors in the weight $\boldsymbol{A}$. This maximisation suggests with the attention weight $\boldsymbol{A}$, we will obtain an inherently explainable decomposition of input patterns.

Overall, the computation in layer $l$ is expressed as 
\begin{equation}
\begin{aligned}
    & \boldsymbol{S}^l = {\rm {E-MHA}}({\rm LN}(\boldsymbol{F}^{l-1})), \\
    & \boldsymbol{Z}^l = \boldsymbol{S}^l + \boldsymbol{F}^{l-1}, \\
    & \boldsymbol{F}^l = {\rm MLP}({\rm LN}(\boldsymbol{Z}^l)) + \boldsymbol{Z}^l,
\end{aligned}
   \label{eq6}
\end{equation}
where ${\rm LN}(\cdot)$ is the LayerNorm layer, ${\rm MLP}(\cdot)$ denotes the multi-layer perceptron layer, and $\boldsymbol{F}^l$ is the output of layer $l$. 

Our key motivation of E-MHA is to dynamically align its attention weights with the discriminative patterns from input values while reducing the impact of noise. The cascade transformer layers in the encoder enable the model to suppress the noise information gradually and learn discriminative input patterns. As a result, the model is able to discover robust representations from the input image. With the attributes of noise robustness and inherent explainability, E-MHA produces the transformer attention map which inherently provides an explainable combination of contributions from discriminative input patterns w.r.t. the model predictions. 

\begin{figure}[t]
  \centering
  \includegraphics[width=1\linewidth, height=0.23\linewidth]{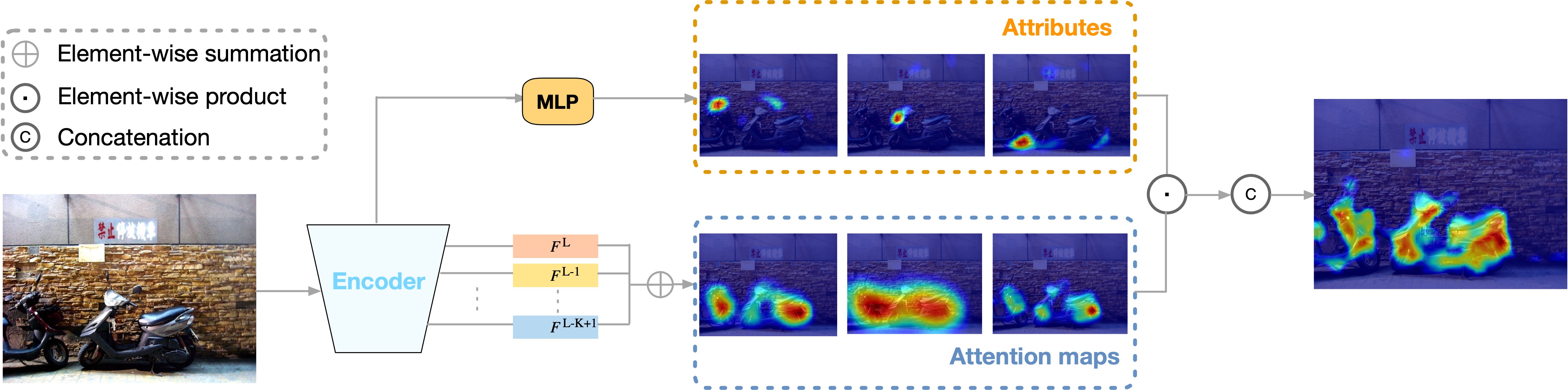}
  \caption{Illustration of Attribute-guided Explainer (AttE). We aggregate the interpretable attention maps from the last $K$ transformer layers to generate a fused attention map with good precision on the complete object context information. The attribute features are regarded as the complement information to better guide the localization of the object context, thus producing robust attribute features in a weakly supervised manner.}
  \label{fig:fig3}
\end{figure}

\subsection{Attribute-guided Explainer (AttE)}
Although the proposed E-MHA provides the intuitive process for explainable feature learning, it is non-trivial to obtain intrinsically interpretable representations that benefit the WSSS tasks without additional regularization. Inspired by the pixel-wise prediction scheme used in semantic segmentation frameworks to localize objects \cite{ShelhamerLD17}, we propose the Attribute-guided Explainer (AttE) module for our eX-ViT with the objective of decomposing the attention map into attribute features based on the diverse attribute discovery. By which, the learned feature maps can be viewed as a set of attributes at a granular level that capture more complete object information. 

Given that the transformer structure tends to learn more uniform representations across all layers, we propose to utilize the transformer attention maps from the last layer in eX-ViT's encoder, to learn a set of trainable attribute features. Concretely, to model the context attention, given an input feature map $\boldsymbol{F} \in \mathbb{R}^{H \times W \times d}$ produced by the encoder $E^{\theta}$'s last layer, we first calculate a set of spatial feature maps that capture the relative importance of all $H ×W$ locations as follows
\begin{equation}
    \boldsymbol{W}_{i,j} = A_\phi(\boldsymbol{F}), \quad \forall \{i, j\} \in H \times W,
    \label{eq7}
\end{equation}
where $A_\phi(\cdot)$ is implemented by a 2-layer MLP block, with one hidden layer followed by a LN layer and the ReLU activation layer. $\boldsymbol{W}_{i,j} \in \mathbb{R}^{c \times H \times W}$ is the obtained embedding with the hidden dimension $c$. We will investigate the influence of $c$ on the model performance in \cref{sec:numberc}.

Furthermore, we apply the $\ell_2$-norm function to $\boldsymbol{W}_{i,j}$ along the channel dimension, which is formally expressed as 
\begin{equation}
    \overline{\boldsymbol{W}}_{i,j} = \frac{\boldsymbol{W}_{i,j}}{||\boldsymbol{W}_{i,j}||_2},
     \label{eq8}
\end{equation}
where $||\cdot||_2$ denotes the L2 norm, $\overline{\boldsymbol{W}}_{i,j}$ is the normalized representation indicating which spatial features to emphasize or suppress.

Subsequently, $\overline{\boldsymbol{W}}$ is sliced into $S$ groups on the channel dimension, i.e., ($\overline{\boldsymbol{W}}_1$, $\overline{\boldsymbol{W}}_2$,..., $\overline{\boldsymbol{W}}_S$), where $\overline{\boldsymbol{W}}_s \in \mathbb{R}^{\frac{c}{S} \times H \times W}$ stands for the feature map of the $s$-th attribute, $S$ is the total number of attributes. To this end, we can apply $\overline{\boldsymbol{W}}_s$ of attribute $s$ to the feature $\boldsymbol{F}$ by 
\begin{equation}
    \boldsymbol{G}_s = \overline{\boldsymbol{W}}_s \odot \boldsymbol{F},
     \label{eq9}
\end{equation}
where $\odot$ is the element-wise product, $\boldsymbol{G}=[\boldsymbol{G}_1,\boldsymbol{G}_2,...,\boldsymbol{G}_S]$ is the final output with attribute features. By this means, each feature map $\boldsymbol{F}$ is projected into $S$ attribute representations that explicitly reveal which pixels are related to the attribute $s$. Likewise, we follow the same procedure described from \cref{eq7} to \cref{eq9}, the attribute representation $\boldsymbol{G}^{'}$ of the second augmented input can be generated accordingly with the momentum encoder $E^{\mathcal{E}}$. And our $AI^{\mathcal{E}}$ is also the exponential moving average of the trained $AI^{\theta}$.

In summary, the output of AttE can be seen as the decomposed contributions for individual attributes. By this means, our model is able to encode semantically explainable features for the target object in an explicit manner, which facilitates the learning of complete object context information. Moreover, we elaborately design our attribute-guided loss function to guide the learning of AttE, which will be presented in next subsection.

\subsection{Attribute-guided loss function}
\label{sec:loss}
A challenging problem for typical vision transformers is that they are not intrinsically interpretable due to lack of the representational power. A recent work, DINO \cite{Caron2021} utilized a self-supervised approach based on vision transformers and concluded that the attentions maps contain features about the semantic information of the image. However, the semantic attributes of the target object were not explored to obtain faithful explanations. In our work, we propose to improve model's interpretability by regularizing its representations with the attribute-guided loss function, i.e., the global-level attribute-guided loss $\mathcal{L}_{\rm global}$, the local-level attribute discriminability loss $\mathcal{L}_{\rm dis}$ loss and the attribute diversity loss $\mathcal{L}_{\rm div}$. On one hand, the $\mathcal{L}_{\rm global}$ encourages the predicted attribute features to approximate the target object, which ensures the faithfulness of the global representations. On the other hand, the $\mathcal{L}_{\rm dis}$ and $\mathcal{L}_{div}$ aim to localize fine-grained attributes through the attribute discriminability mechanism and attribute diversity mechanism, thus enabling the robust feature learning.

Since higher layers discover high-level concepts like objects or scenes, we propose to fuse transformer attention maps from the last $K$ encoder layers to achieve good accuracy on the complete object context information. Hence, given the obtained attention map $\boldsymbol{F}^l$ in $l$-th encoder layer, the fused attention map is expressed as
\begin{equation}
    \hat{\boldsymbol{F}} = \frac{1}{K} \sum_l^K \boldsymbol{F}^l,
    \label{eq10}
\end{equation}
where $\hat{\boldsymbol{F}}$ is the fused transformer attention map. By this means, we aggregate cascaded encoder blocks to produce reliable attention map for the complete object localization. As the aggregated attention map  $\hat{\boldsymbol{F}}$ is attribute-agnostic, we propose to couple it with the attribute features $\boldsymbol{G}$ to generate the attribute-guided attention map. The process is defined as follows
\begin{equation}
    \boldsymbol{M} = \hat{\boldsymbol{F}} \odot \boldsymbol{G},
    \label{eq11}
\end{equation}
where $\boldsymbol{M}$ represents the final output of the attribute-guided feature map. 

Based upon $\boldsymbol{M}$, the global-level attribute-guided loss $\mathcal{L}_{\rm global}$ is computed by the multi-label soft margin loss
\begin{equation}
\mathcal{L}_{\rm global} = \frac{1}{C}\sum_{c=1}^C(y^c \log(\hat{y}_c) + (1-y^c)\log(1-\hat{y}_c)),
    \label{eq12}
\end{equation}
where the prediction $\hat{y}_c$ is obtained by feeding the feature map $\boldsymbol{M}$ into a classification layer. By optimizing the $\mathcal{L}_{\rm global}$, the interpretable features are gathered as a summation of the important scores of all attribute features, which ensures the faithfulness of the explanations.

In addition, in order to improve the ability of network to learn diverse and discriminative attribute representations for the target object, we propose the local-level attribute-guided loss through the attribute discriminability mechanism and attribute diversity mechanism in a self-supervised manner. Intuitively, the attribute discriminability mechanism aims to make attribute features consistent discriminative between two types of input views, while the attribute diversity mechanism enables the model to learn the effective decomposition with the attribute diversity. Formally, the attribute discriminability loss $\mathcal{L}_{\rm dis}$ is defined by
\begin{equation}
    \begin{aligned}
    & \mathcal{L}_{\rm dis} = |d-\sum_{s=1}^S d^s|, \\
    & d = \ell(g(\boldsymbol{G}), g(\boldsymbol{G}^{'})), \\
    & d_s = \ell(g(\boldsymbol{G}_s), g(\boldsymbol{G}_s^{'})),
    \end{aligned}
    \label{eq13}
\end{equation}
where $g(\cdot)$ is the generalized mean pooling. And we adopt the normalized Mean Square Error in BYOL \cite{GrillSATRBDPGAP20} as the $\ell(\cdot)$ function to calculate distance between two features. As can be seen from \cref{eq13}, $d$ is leveraged to minimize the difference between attribute features, while $d_s$ is used to guarantee the consistency between $\boldsymbol{G}$ and $\boldsymbol{G}^{'}$ for each individual attribute. Empirically, this attribute discriminability loss function $\mathcal{L}_{\rm dis}$ is able to facilitate the model to discover discriminative class-specific attributes and obtain more comprehensive localization maps. Meanwhile, we introduce the attribute diversity loss $\mathcal{L}_{\rm div}$ is formally defined by
\begin{equation}
    \mathcal{L}_{\rm div} = \frac{1}{S(S-1)} \sum_{i=1,j=1}^S \sum_{i \neq j}^S \frac{<\boldsymbol{G}_i, \boldsymbol{G}_j>}{||\boldsymbol{G}_i||_2 ||\boldsymbol{G}_j||_2},
    \label{eq14}
\end{equation}
The intuition behind the $\mathcal{L}_{\rm div}$ is to make attribute features to the maximally independent from each other, so as to make attribute features focus on different discriminative object regions.

Overall, the loss function for the proposed eX-ViT is given below
\begin{equation}
    \mathcal{L} = \mathcal{L}_{\rm global} + \alpha \mathcal{L}_{\rm dis} + \beta \mathcal{L}_{\rm div},
    \label{eq15}
\end{equation}
where $\mathcal{L}_{\rm global}$ is the multi-label soft margin loss. $\alpha$ and $\beta$ are the coefficient of $\mathcal{L}_{\rm dis}$ and $\mathcal{L}_{\rm div}$, respectively.

As a result, our attribute-guided loss promotes the learning of attribute  features. The global-level loss $\mathcal{L}_{\rm global}$ ensures a faithful transformer model, while the $\mathcal{L}_{\rm dis}$ and $\mathcal{L}_{\rm div}$ enable discriminative and robust attribute features. The effectiveness of the loss function is further verified in the experimental section.

\newcommand{\tabincell}[2]{\begin{tabular}{@{}#1@{}}#2\end{tabular}} 

\section{Experiments}
\label{sec:experi}
In this section, we first introduce the experimental settings including datasets and implementation details. Second, we evaluate the efficiency of our proposed eX-ViT and compare it with the recent state-of-the-art methods. Third, we conduct a series of ablation studies to discover the performance contribution from different modules in our framework.

\subsection{Setup}
\subsubsection{Datasets}
We conduct experiments on PASCAL VOC 2012 dataset \cite{EveringhamEGWWZ15} and MS COCO 2014 dataset \cite{lin2014microsoft}. PASCAL VOC 2012 dataset includes 20 object classes and one background class for the semantic segmentation task. Following the common experimental configuration from SBD \cite{HariharanABMM11}, we adopt the augmented dataset which contains three subsets, training, validation, and testing sets, each having 10582, 1449, and 1464 images, respectively. MS COCO 2014 dataset uses 81 classes, its training and validation sets have 82081 images and 40137 images respectively. Note image-level labels are only used during training and ground-truth bounding box annotations are solely used during test. In line with previous works \cite{ruaffinity2022}, we report the mean Intersection-over-Union (mIoU) to evaluate the performance of our proposed model.

\subsubsection{Implementation details}
We use PyTorch for implementation and conduct experiments. The encoder parameters are pre-trained on ImageNet \cite{russakovsky2015imagenet}, while other parameters are initialized as suggested in \cite{HeZR015}. During training, we use the AdamW optimizer. For the transformer encoder $E^{\theta}$, the initial learning rate is set to be $5 \times 10^{-5}$, which is further decayed via a polynomial schedule. The learning rate for the rest of the parameters is $5 \times 10^{-4}$. For the training on the PASCAL VOC 2012 dataset, the batch size is set as 16, and the training process lasts 40k iterations. On MS COCO 2014 dataset, we trained the models for 80k iterations with a batch size of 8. For data augmentation, we used random scaling with a range of [0.5,2.0], random horizontally flipping, and random cropping. 

The default hyper-parameters are set as follows. For encoders $E^{\theta}$ and $E^{\mathcal{E}}$, it contains 12 layers, 6 heads within each E-MHA, and the hidden dimension is set to 384. Empirically, we set $\alpha$ and $\beta$ in \cref{eq15} as 0.5 and 1.0 respectively throughout this paper. In line with previous works, we use the ResNet38 \cite{WuSH19} as the backbone for semantic segmentation. At test time, since our eX-ViT is a siamese network, only one branch is needed. Following the common practice in prior stuides \cite{WangZKSC20}, we use multi-scale testing and CRFs to obtain pseudo segmentation results.

\begin{table}
  \caption{mIoU (\%) of localization maps on the PASCAL VOC 2012 training set.}
  \centering
  \setlength{\tabcolsep}{2pt}
  \begin{tabular}{l|c|c}
  \toprule  % 顶部线 
  Method & Local. Maps & +denseCRF \\
  \midrule  % 中部线
  (CVPR'20) SCE \cite{ChangWHPT020} & 50.9 & 55.3 \\
  (CVPR'20) SEAM \cite{WangZKSC20} & 55.4 & 56.8 \\
  (CVPR'21) EDAM \cite{WuHGWWLL21} & 52.8 & 58.2 \\
  (CVPR'21) AdvCAM \cite{LeeKY21} & 55.6 & 62.1 \\
  (ICCV'21) ECS-Net \cite{SunSZH21} & 56.6 & 58.6 \\
  (ICCV'21) CSE \cite{KweonYKPY21} & 56.0 & 62.8 \\
  (CVPR'22) SIPE \cite{Chen02909} & 58.6 & 64.7 \\
  (CVPR'22) ReCAM \cite{chencrea962} & 56.6 & - \\
  \bf{(Ours) eX-ViT} & \bf{59.1} & \bf{65.3} \\
  \bottomrule
  \end{tabular}
\label{table:tab1}
\end{table}

\subsection{Comparison with state-of-the-arts}
\subsubsection{Comparison on localization maps}
\begin{figure}[htbp]
  \centering
  \setlength{\abovecaptionskip}{0.cm}
    \subfloat{
        \begin{minipage}[b]{0.03\textwidth}
     \rotatebox{90}{~~~~\scalebox{1.2}{Image}}
    \rotatebox{90}{~~~~~\scalebox{1}[1.2]{Ground-truth}}
   \rotatebox{90}{~~~~~~~~~~\scalebox{1.2}[1.2]{CAM}}
    \rotatebox{90}{~~~~~~~\scalebox{1.2}[1.2]{SIPE}}
    \rotatebox{90}{~~~~~\scalebox{1.2}[1.2]{AdvCAM}}
    \rotatebox{90}{~~~~~~~~\scalebox{1.2}[1.2]{eX-ViT}}
    \label{fig:short-4a}
        \end{minipage}}
        \hspace{-5pt}
  \subfloat{
        \begin{minipage}[b]{0.18\textwidth}
     \includegraphics[width=1\textwidth,height=0.8\textwidth]{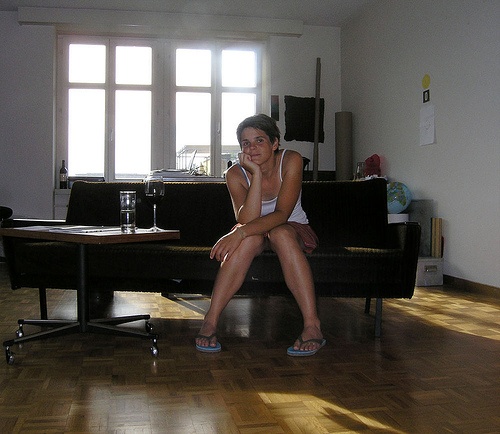}
    \includegraphics[width=1\linewidth, height=0.8\linewidth]{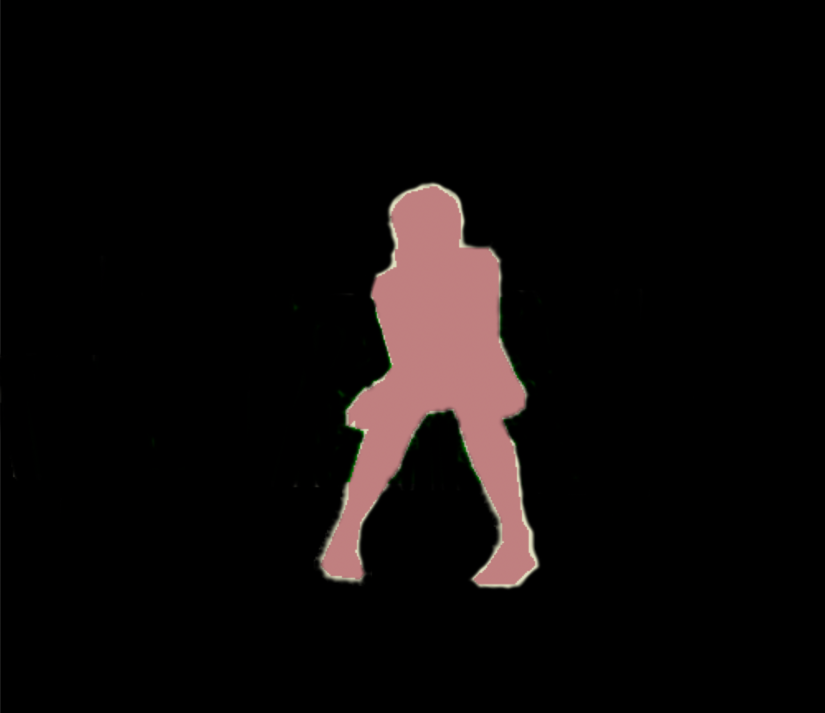}
   \includegraphics[width=1\linewidth, height=0.8\linewidth]{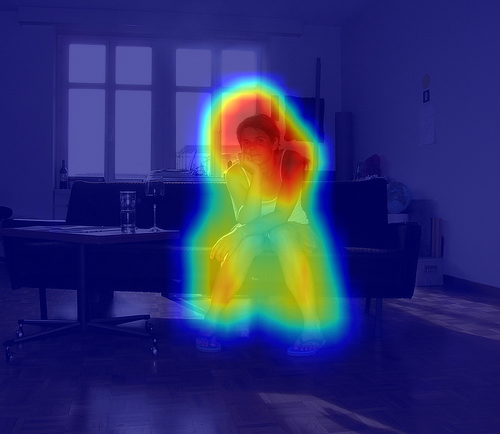}
    \includegraphics[width=1\linewidth, height=0.8\linewidth]{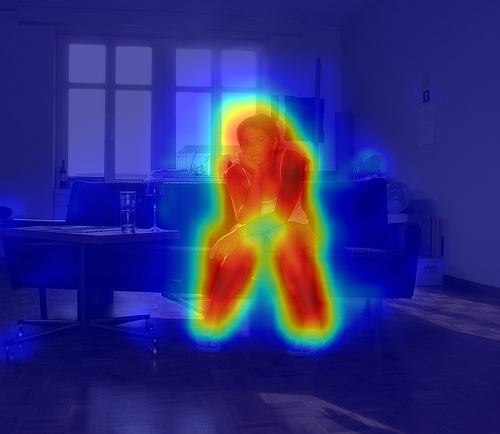}
    \includegraphics[width=1\linewidth, height=0.8\linewidth]{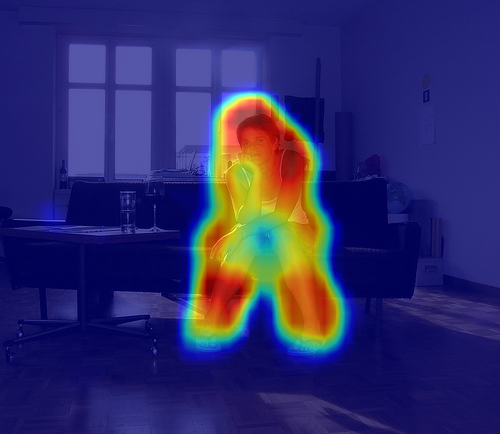}
    \includegraphics[width=1\linewidth, height=0.8\linewidth]{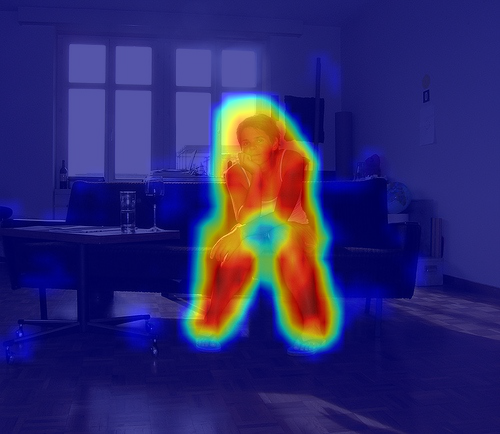}
    \label{fig:short-4b}
        \end{minipage}}
        \hspace{-5pt}
    \subfloat{
        \begin{minipage}[b]{0.18\textwidth}
    \includegraphics[width=1\textwidth,height=0.8\textwidth]{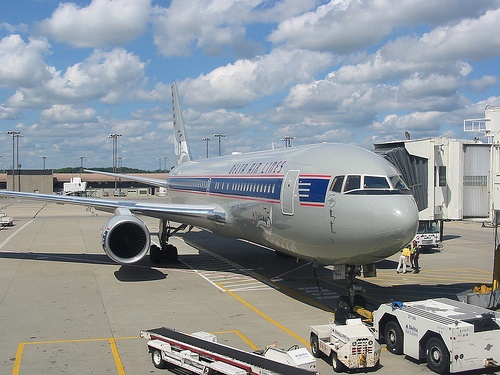}
    \includegraphics[width=1\textwidth,height=0.8\textwidth]{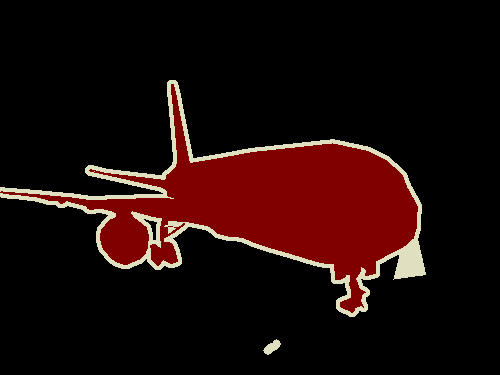}
    \includegraphics[width=1\textwidth,height=0.8\textwidth]{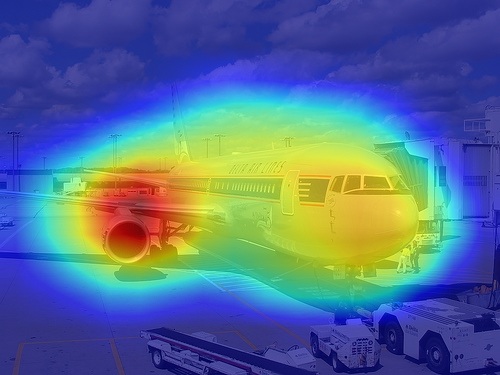}
    \includegraphics[width=1\textwidth,height=0.8\textwidth]{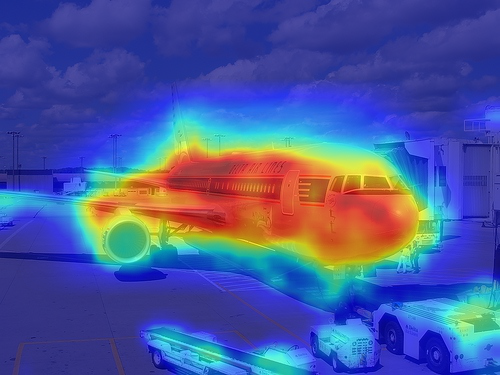}
    \includegraphics[width=1\textwidth,height=0.8\textwidth]{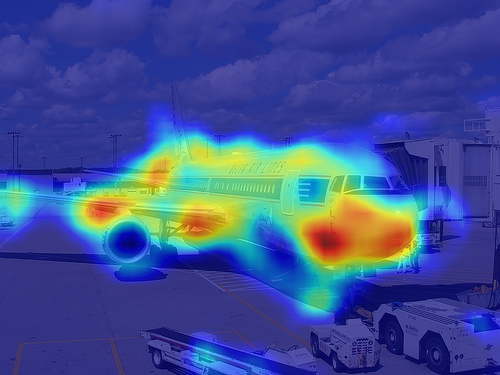}
    \includegraphics[width=1\textwidth,height=0.8\textwidth]{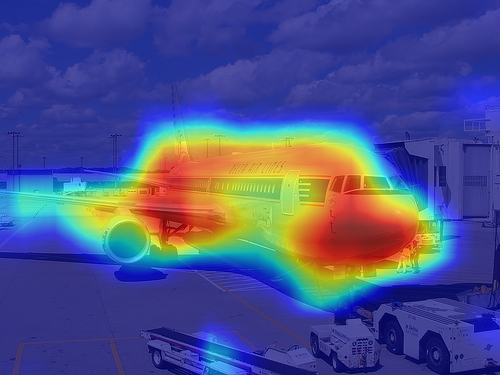}
    \label{fig:short-4c}
        \end{minipage}}
        \hspace{-5pt}
    \subfloat{
        \begin{minipage}[b]{0.18\textwidth}
    \includegraphics[width=1\textwidth,height=0.8\textwidth]{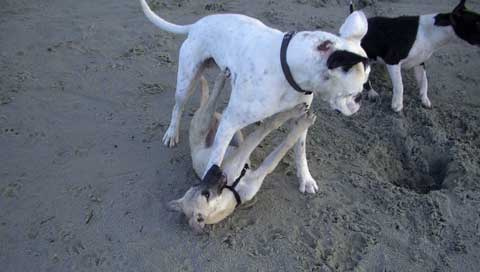}
    \includegraphics[width=1\textwidth,height=0.8\textwidth]{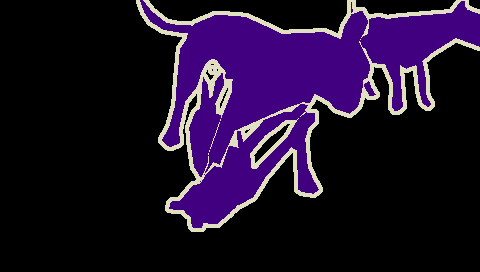}
    \includegraphics[width=1\textwidth,height=0.8\textwidth]{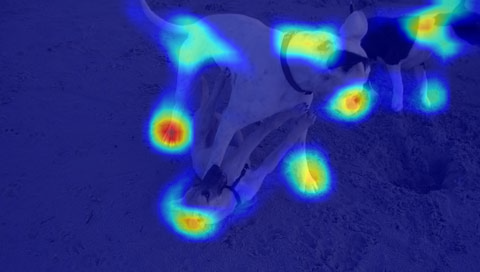}
    \includegraphics[width=1\textwidth,height=0.8\textwidth]{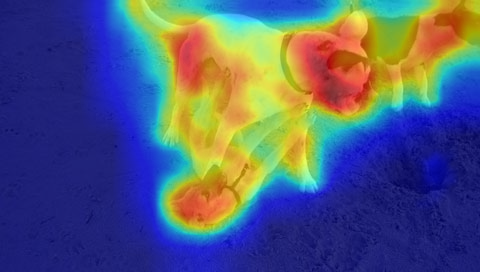}
    \includegraphics[width=1\textwidth,height=0.8\textwidth]{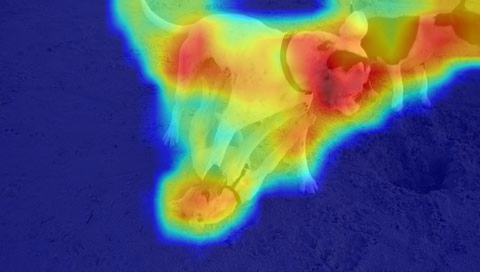}
    \includegraphics[width=1\textwidth,height=0.8\textwidth]{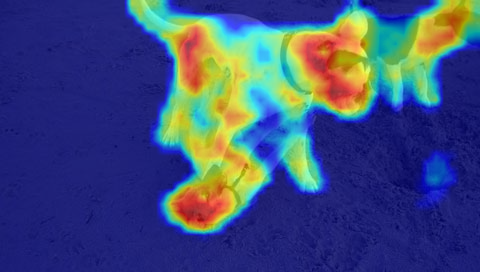}
    \label{fig:short-4d}
        \end{minipage}}
        \hspace{-5pt}
    \subfloat{
        \begin{minipage}[b]{0.18\textwidth}
    \includegraphics[width=1\textwidth,height=0.8\textwidth]{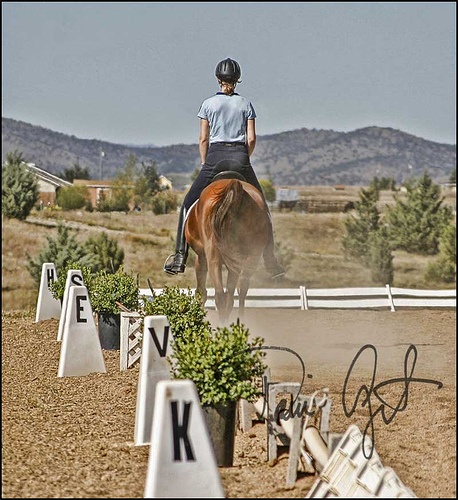}
    \includegraphics[width=1\textwidth,height=0.8\textwidth]{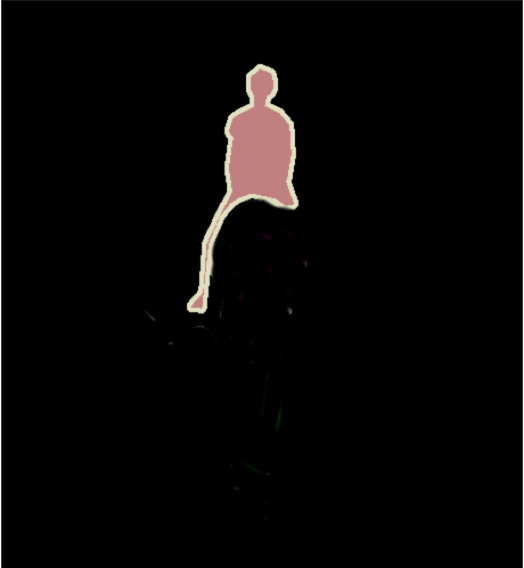}
    \includegraphics[width=1\textwidth,height=0.8\textwidth]{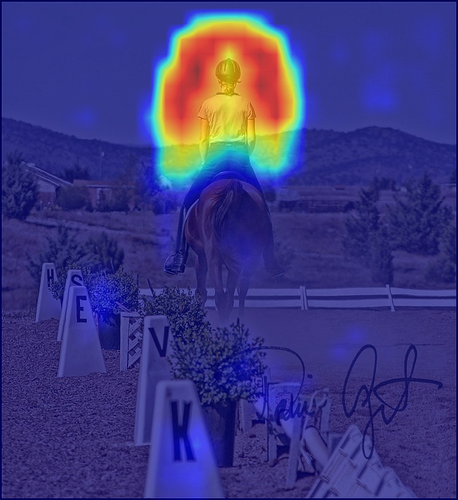}
    \includegraphics[width=1\textwidth,height=0.8\textwidth]{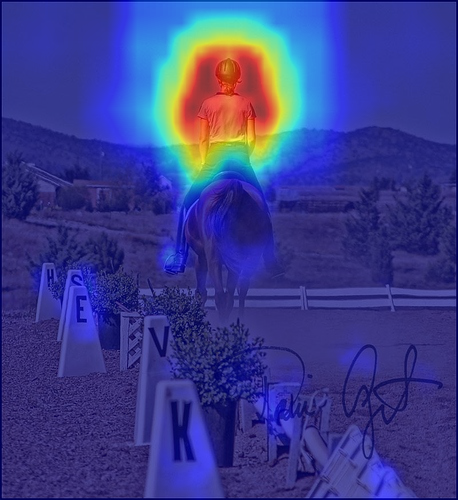}
    \includegraphics[width=1\textwidth,height=0.8\textwidth]{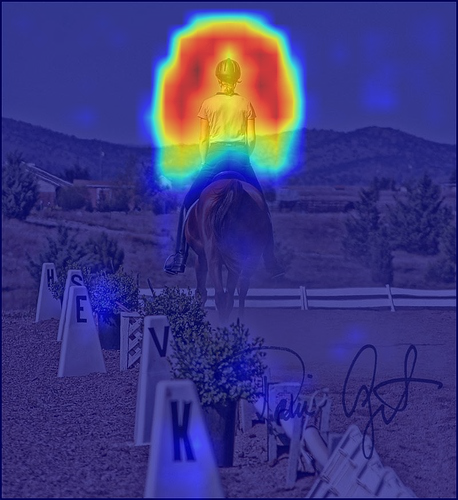}
    \includegraphics[width=1\textwidth,height=0.8\textwidth]{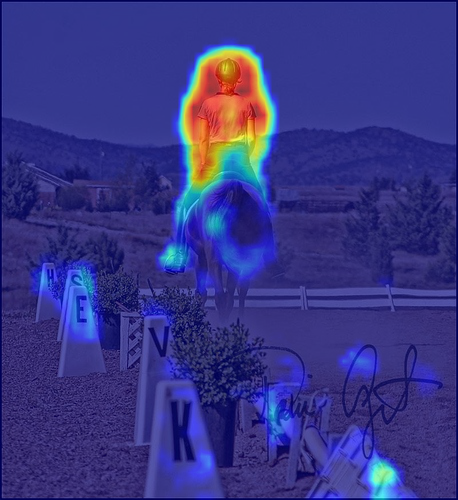}
    \label{fig:short-4e}
        \end{minipage}}
    \hspace{-5pt}
    \subfloat{
        \begin{minipage}[b]{0.18\textwidth}
    \includegraphics[width=1\textwidth,height=0.8\textwidth]{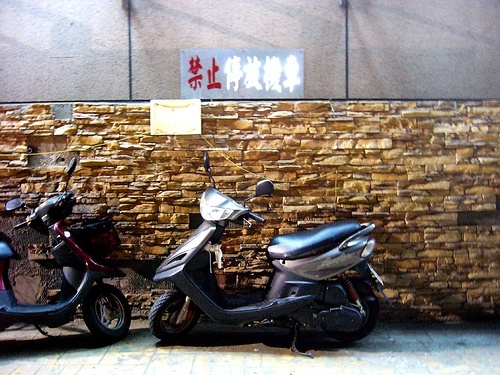}
    \includegraphics[width=1\textwidth,height=0.8\textwidth]{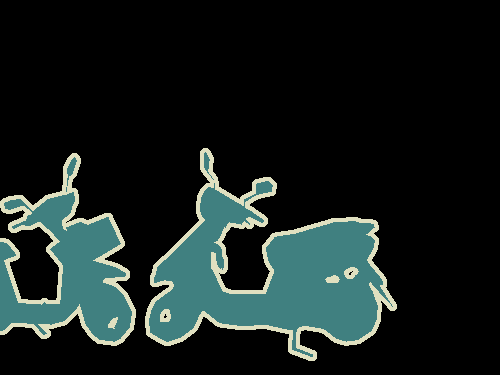}
    \includegraphics[width=1\textwidth,height=0.8\textwidth]{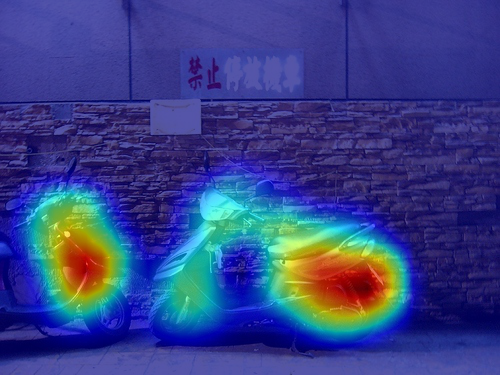}
    \includegraphics[width=1\textwidth,height=0.8\textwidth]{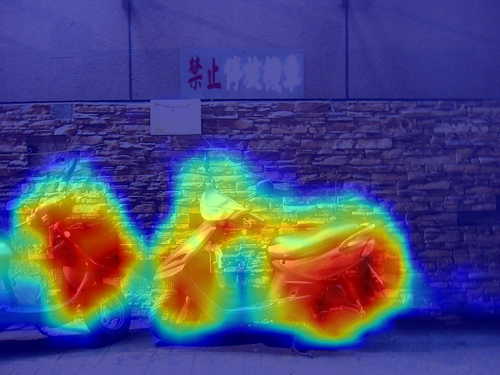}
    \includegraphics[width=1\textwidth,height=0.8\textwidth]{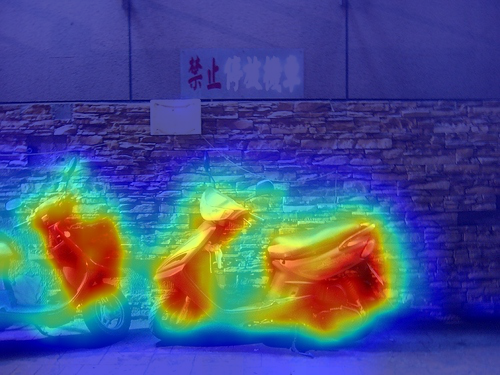}
    \includegraphics[width=1\textwidth,height=0.8\textwidth]{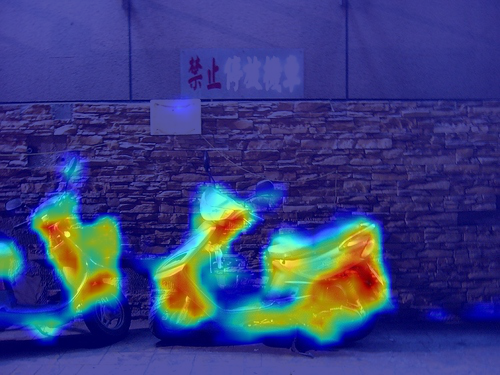}
    \label{fig:short-4f}
        \end{minipage}}
\caption{Visual comparison of localization maps generated by different methods on PASCAL VOC 2012 training set. From top to down: original image, ground-truth, CAM \cite{zhou2016learning}, SIPE \cite{Chen02909}, AdvCAM \cite{LeeKY21} and our eX-ViT.}
\label{fig:fig4}
\end{figure}

We first evaluate the qualitative results of mIoU on localization maps, where those pseudo semantic segmentation results are obtained by applying PSA \cite{AhnK18} on the localization maps produced by our proposed eX-ViT. \cref{table:tab1} reports the results of our proposed method as well as other recent state-of-the-art approaches on the PASCAL VOC 2012 training set. As can be seen from the table, SIPE \cite{Chen02909} achieves the state-of-the-art result with a mIoU of 58.6\%. eX-ViT outperforms all compared methods in terms of both metrics. Concretely, the results show that our eX-ViT improves the mIoU to 59.1\%. We also conduct experiments based on eX-ViT with denseCRF post-processing, and the gain becomes up to 65.3\%. \cref{fig:fig4} shows visual comparison of object localization maps on PASCAL VOC 2012 training set. As shown in \cref{fig:fig4}, the fused class-specific attribute-guided localization map can effectively capture the discriminative features within the object context of the target class with more useful clues. As a result, the fused localization map by use of our eX-ViT brings notable visual improvements to produce complete localization maps.

\begin{table}
  \caption{Performance comparison of state-of-the-art methods in terms of mIoU (\%) on the PASCAL VOC 2012 validation and test sets. $Sup.$ indicates supervision type. $\mathcal{F}$: full supervision; $\mathcal{I}$: image-level labels; $\mathcal{S}$: saliency maps.}
  \centering
  \setlength{\tabcolsep}{2pt}
  \begin{tabular}{l|l|l|cc}
  \toprule  % 顶部线 
  Method & \makebox[0.1\textwidth][l]{$Sup.$} & \makebox[0.1\textwidth][l]{Backbone} & \makebox[0.1\textwidth][c]{validation} & \makebox[0.1\textwidth][c]{test} \\
  \midrule  % 中部线
  \multicolumn{5}{l}{\bf{Fully-supervised methods}} \\
  (TPAMI'18) DeepLab \cite{ChenPKMY18} & $\mathcal{F}$ & ResNet101 & 77.6 & 79.7 \\
  (PR'19) WideResNet38 \cite{WuSH19} & $\mathcal{F}$ & WR38 & 80.8 & 82.5 \\
  (NeurIPS'21) Segformer \cite{Xie15203} & $\mathcal{F}$ & MiT-B1 & 78.7 & - \\
  \hline
  \multicolumn{5}{l}{\bf{Weakly-supervised methods}} \\
  (ECCV'20) MCIS \cite{SunWDG20} & $\mathcal{I} + \mathcal{S}$ & ResNet101 & 66.2 & 66.9 \\
  (ICCV'21) AuxSegNet \cite{XuOBBS021} & $\mathcal{I} + \mathcal{S}$ & ResNet38 & 69.0 & 68.6 \\
  (CVPR'21) NSROM \cite{Yao0XZS0T021} & $\mathcal{I} + \mathcal{S}$ & ResNet101 & 70.4 & 70.2 \\
  (CVPR'21) EPS \cite{LeeLLS21} & $\mathcal{I} + \mathcal{S}$ & ResNet101 & 70.9 & 70.8 \\
  (CVPR'21) EDAM \cite{WuHGWWLL21} & $\mathcal{I} + \mathcal{S}$ & ResNet101 & 70.9 & 70.6 \\
  (ICCV'19) OAA+ \cite{JiangHCCWX19} & $\mathcal{I} + \mathcal{S}$ & ResNet101 & 65.2 & 66.4 \\
  (TPAMI'21) A2GNN \cite{Zhang04054} & $\mathcal{I}$ & ResNet101 & 66.8 & 67.4 \\
  (TPAMI'22) LIID \cite{LiuWWSQC22} & $\mathcal{I}$ & ResNet101 & 66.5 & 67.5 \\
  (CVPR'20) SEAM \cite{WangZKSC20} & $\mathcal{I}$ & ResNet38 & 64.5 & 65.7 \\
  (AAAI'20) RRM \cite{ZhangXWSH20} & $\mathcal{I}$ & ResNet38 & 62.6 & 62.9 \\
  (NeurIPS'20) CONTA \cite{ZhangZT0S20} & $\mathcal{I}$ & ResNet38 & 66.1 & 66.7 \\
  (ICCV'21) CDA \cite{SuSLW21} & $\mathcal{I}$ & ResNet38 & 66.1 & 66.8 \\
  (ICCV'21) ECS-Net \cite{SunSZH21} & $\mathcal{I}$ & ResNet38 & 66.6 & 67.6 \\
  (ICCV'21) CSE \cite{KweonYKPY21} & $\mathcal{I}$ & ResNet38 & 68.4 & 68.2 \\
  (CVPR'21) AdvCAM \cite{LeeKY21} & $\mathcal{I}$ & ResNet101 & 68.1 & 68.0 \\
  (NeurIPS'21) RIB \cite{Lee06530} & $\mathcal{I}$ & ResNet101 & 68.3 & 68.6 \\
  (CVPR'22) SIPE \cite{Chen02909} & $\mathcal{I}$ & ResNet101 & 68.8 & 69.7 \\
  (CVPR'22) ReCAM \cite{chencrea962} & $\mathcal{I}$ & ResNet101 & 68.5 & 68.4 \\
  (CVPR'22) Ru et al. \cite{ruaffinity2022}  & $\mathcal{I}$ & MiT-B1 & 66.0 & 66.3 \\
  \bf{(Ours) eX-ViT} & $\mathcal{I}$ & ResNet38 & \bf{71.2} & \bf{71.1} \\
  \bottomrule
  \end{tabular}
\label{table:tab2}
\end{table}

\begin{figure}
  \centering
  \setlength{\abovecaptionskip}{0.cm}
      \subfloat{
        \begin{minipage}[b]{0.03\textwidth}
     \rotatebox{90}{~~~~~~~~\scalebox{1.2}{Image}}
    \rotatebox{90}{~~~~~~~~~\scalebox{1}[1.2]{Ground-truth}}
    \rotatebox{90}{~~~~~~~~~~~~\scalebox{1.2}[1.2]{SIPE}}
    \rotatebox{90}{~~~~~~~~~~~\scalebox{1.2}[1.2]{eX-ViT}}
    \label{fig:short-5a}
        \end{minipage}}
        \hspace{-5pt}
     \subfloat{
        \begin{minipage}[b]{0.22\textwidth}
     \includegraphics[width=1\textwidth,height=0.8\textwidth]{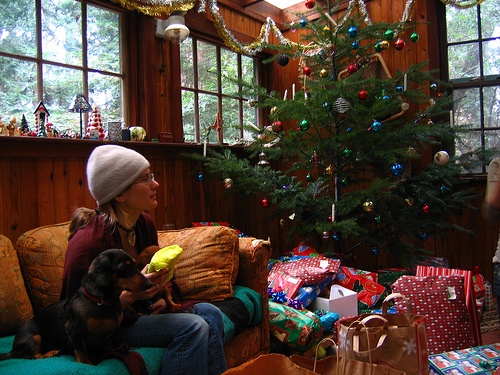}
     \includegraphics[width=1\linewidth, height=0.8\linewidth]{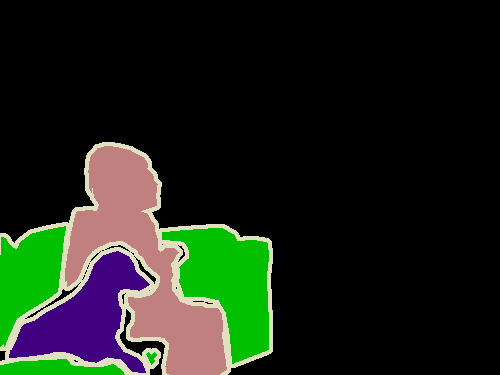}
     \includegraphics[width=1\linewidth, height=0.8\linewidth]{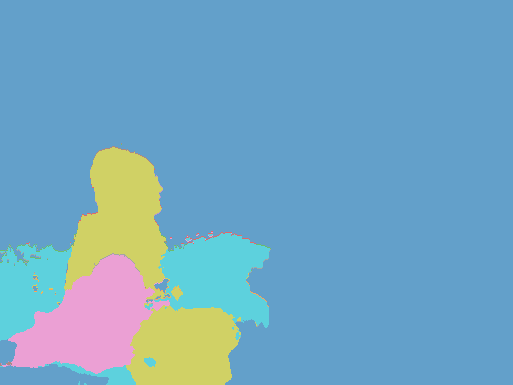}
     \includegraphics[width=1\linewidth, height=0.8\linewidth]{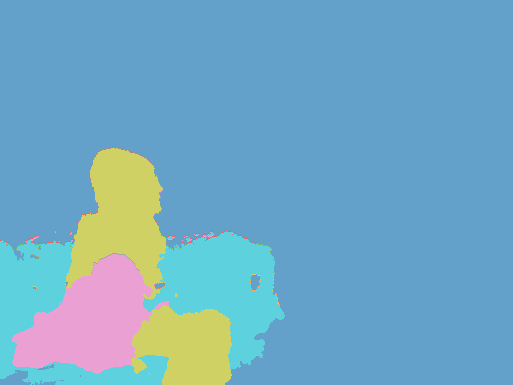}
    \label{fig:short-5b}
        \end{minipage}}
        \hspace{-5pt}
    \subfloat{
        \begin{minipage}[b]{0.22\textwidth}
    \includegraphics[width=1\textwidth,height=0.8\textwidth]{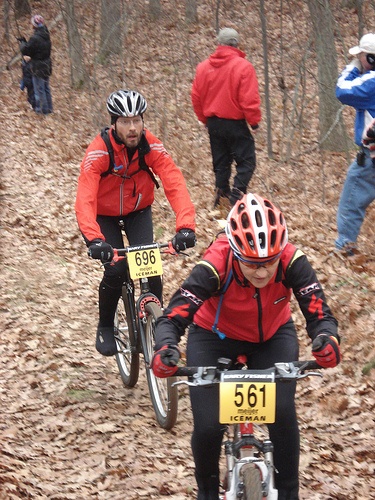}
    \includegraphics[width=1\textwidth,height=0.8\textwidth]{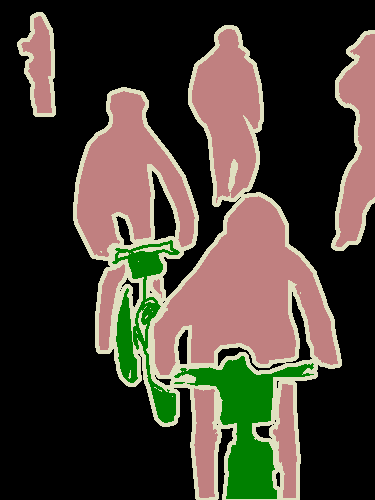}
    \includegraphics[width=1\textwidth,height=0.8\textwidth]{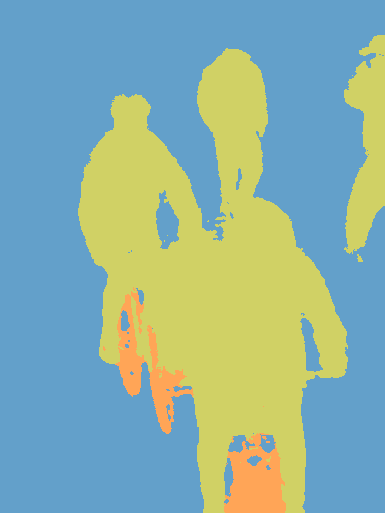}
    \includegraphics[width=1\textwidth,height=0.8\textwidth]{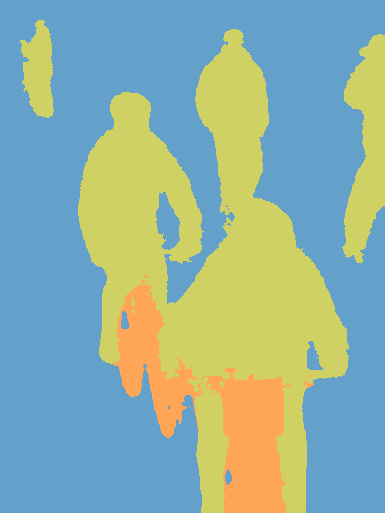}
    \label{fig:short-5c}
        \end{minipage}}
        \hspace{-5pt}
    \subfloat{
        \begin{minipage}[b]{0.22\textwidth}
    \includegraphics[width=1\textwidth,height=0.8\textwidth]{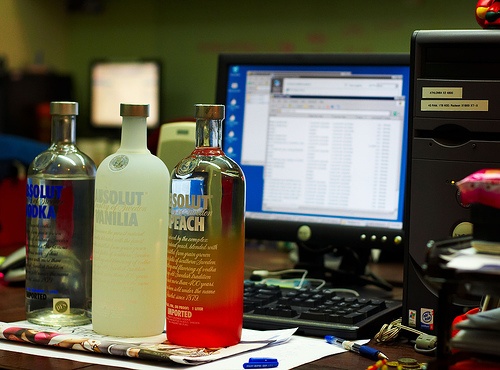}
    \includegraphics[width=1\textwidth,height=0.8\textwidth]{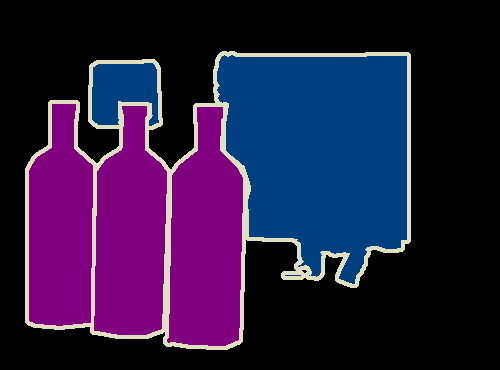}
    \includegraphics[width=1\textwidth,height=0.8\textwidth]{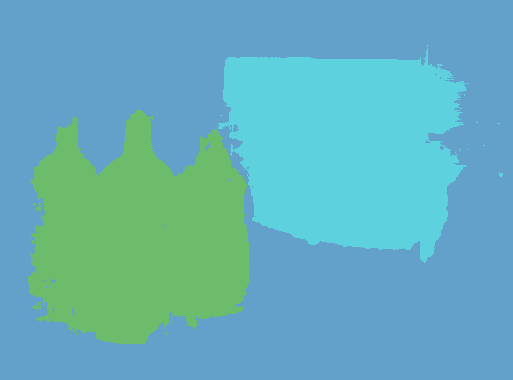}
    \includegraphics[width=1\textwidth,height=0.8\textwidth]{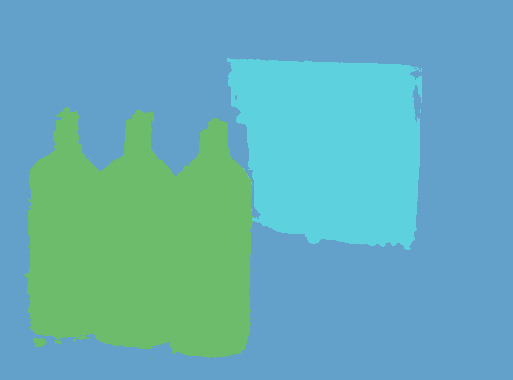}
    \label{fig:short-5d}
        \end{minipage}}
        \hspace{-5pt}
    \subfloat{
        \begin{minipage}[b]{0.22\textwidth}
    \includegraphics[width=1\textwidth,height=0.8\textwidth]{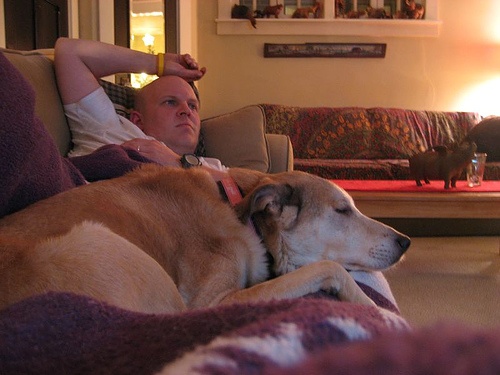}
    \includegraphics[width=1\textwidth,height=0.8\textwidth]{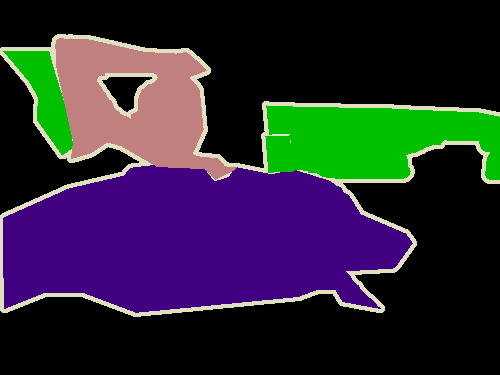}
    \includegraphics[width=1\textwidth,height=0.8\textwidth]{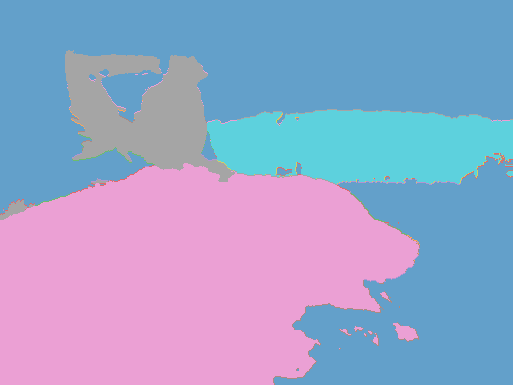}
    \includegraphics[width=1\textwidth,height=0.8\textwidth]{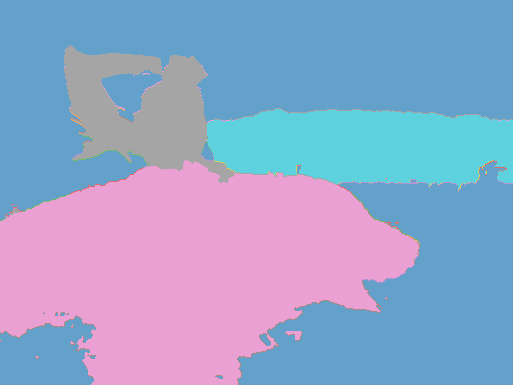}
    \label{fig:short-5e}
        \end{minipage}}
\caption{Qualitative segmentation results on the validation set of PASCAL VOC 2012. From top to down: original image, ground-truth, SIPE \cite{Chen02909} and our eX-ViT.}
\label{fig:fig5}
\end{figure}

\begin{figure}[t]
  \centering
  \setlength{\abovecaptionskip}{0.cm}
    \subfloat{
        \begin{minipage}[b]{0.03\textwidth}
     \rotatebox{90}{~~~~~~~~~~\scalebox{1.2}{Image}}
    \rotatebox{90}{~~~~~~~~~\scalebox{1}[1.2]{Ground-truth}}
    \rotatebox{90}{~~~~~~~~~~~~\scalebox{1.2}[1.2]{SIPE}}
    \rotatebox{90}{~~~~~\scalebox{1.2}[1.2]{eX-ViT}}
    \label{fig:short-6a}
        \end{minipage}}
        \hspace{-5pt}
    \subfloat{
        \begin{minipage}[b]{0.22\textwidth}
     \includegraphics[width=0.9\textwidth,height=0.8\textwidth]{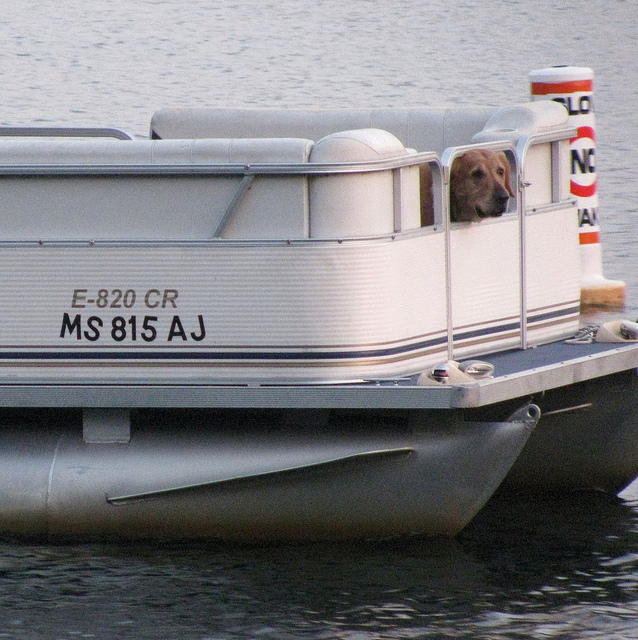}
     \includegraphics[width=0.9\linewidth, height=0.8\linewidth]{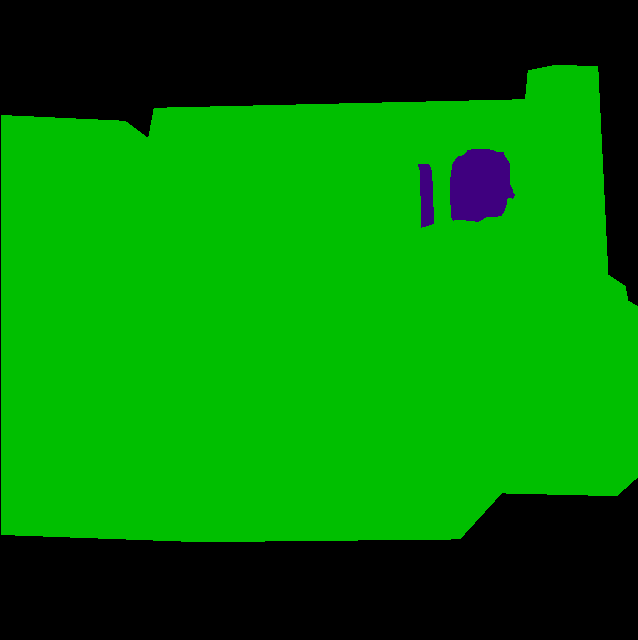}
     \includegraphics[width=0.9\linewidth, height=0.8\linewidth]{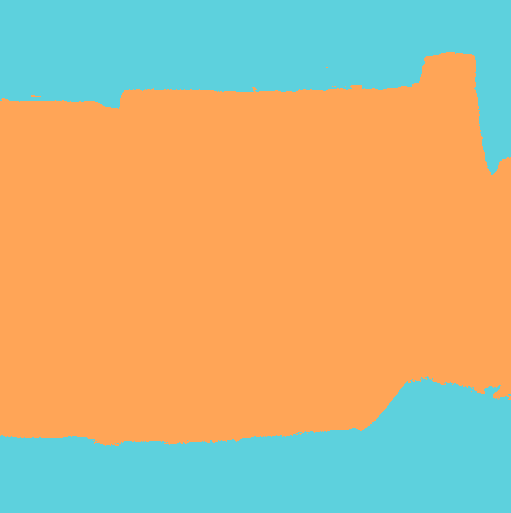}
     \includegraphics[width=0.9\linewidth, height=0.8\linewidth]{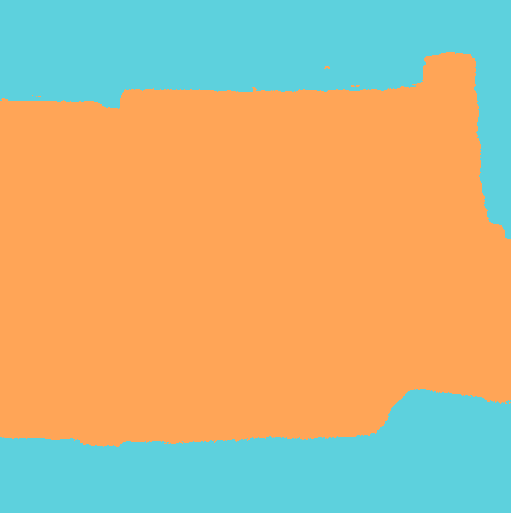}
    \label{fig:short-6b}
        \end{minipage}}
        \hspace{-14pt}
    \subfloat{
        \begin{minipage}[b]{0.22\textwidth}
    \includegraphics[width=0.9\textwidth,height=0.8\textwidth]{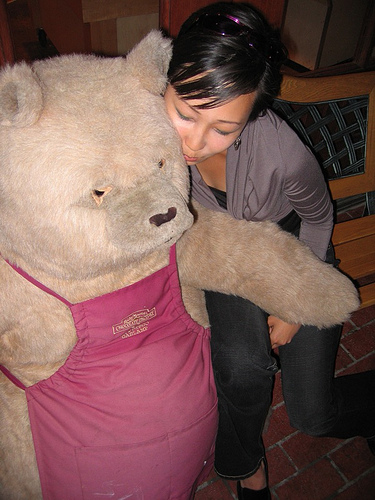}
    \includegraphics[width=0.9\textwidth,height=0.8\textwidth]{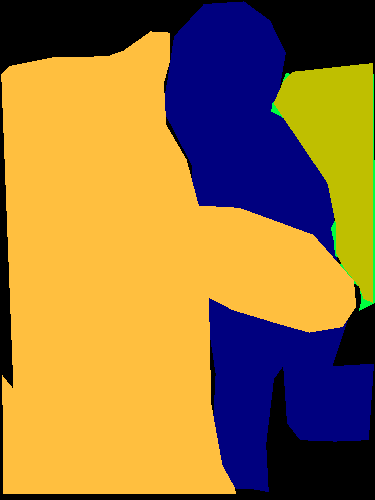}
    \includegraphics[width=0.9\textwidth,height=0.8\textwidth]{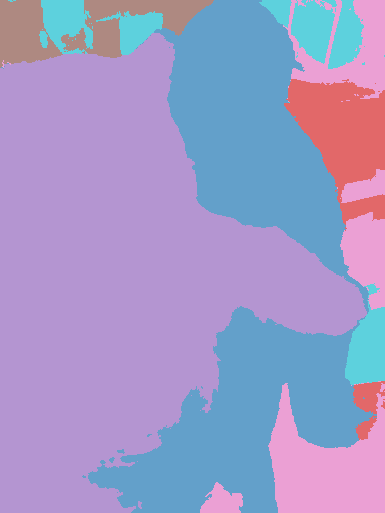}
    \includegraphics[width=0.9\textwidth,height=0.8\textwidth]{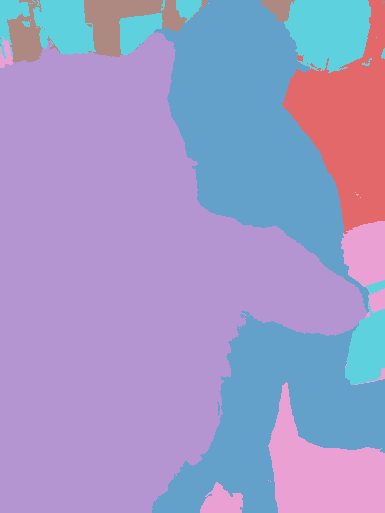}
    \label{fig:short-6c}
        \end{minipage}}
        \hspace{-14pt}
    \subfloat{
        \begin{minipage}[b]{0.22\textwidth}
    \includegraphics[width=0.9\textwidth,height=0.8\textwidth]{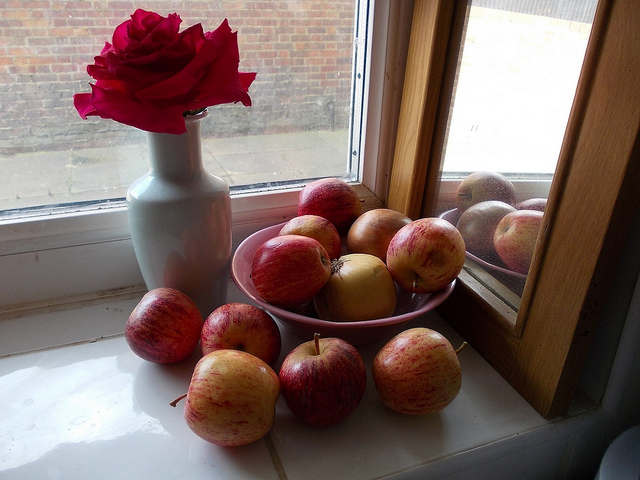}
    \includegraphics[width=0.9\textwidth,height=0.8\textwidth]{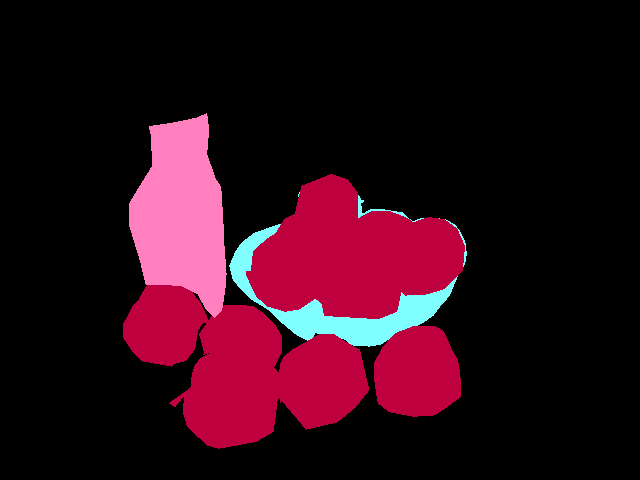}
    \includegraphics[width=0.9\textwidth,height=0.8\textwidth]{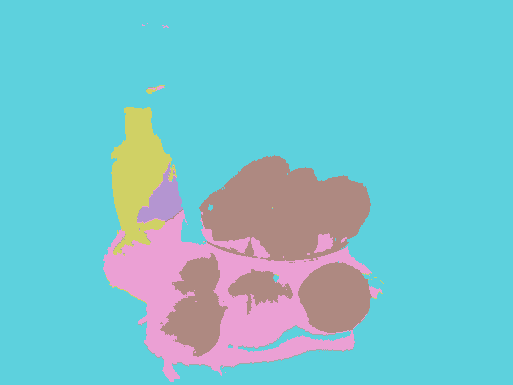}
    \includegraphics[width=0.9\textwidth,height=0.8\textwidth]{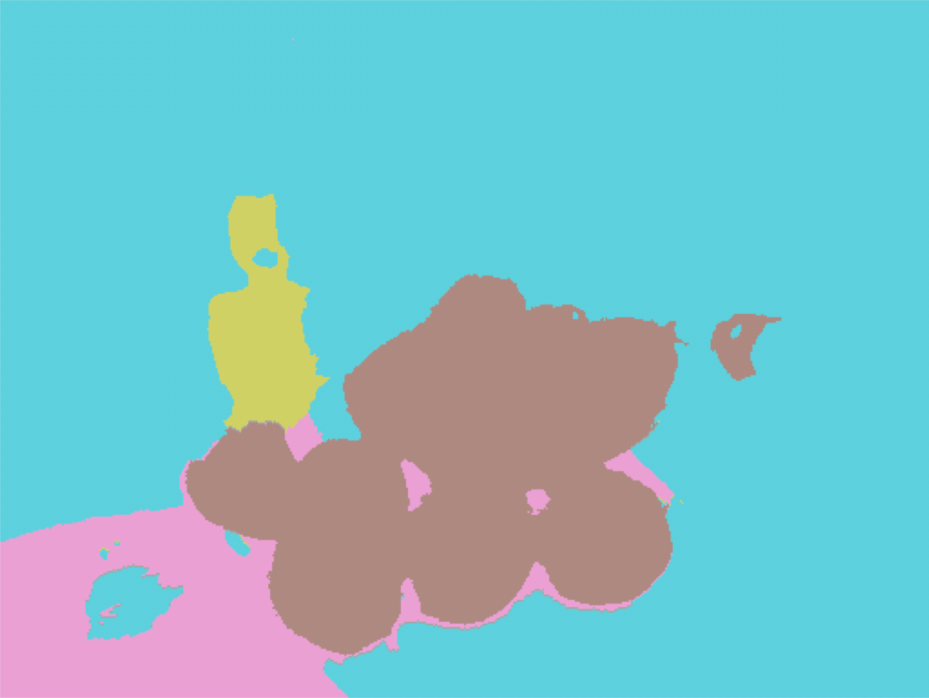}
    \label{fig:short-6d}
        \end{minipage}}
        \hspace{-14pt}
    \subfloat{
        \begin{minipage}[b]{0.22\textwidth}
    \includegraphics[width=0.9\textwidth,height=0.8\textwidth]{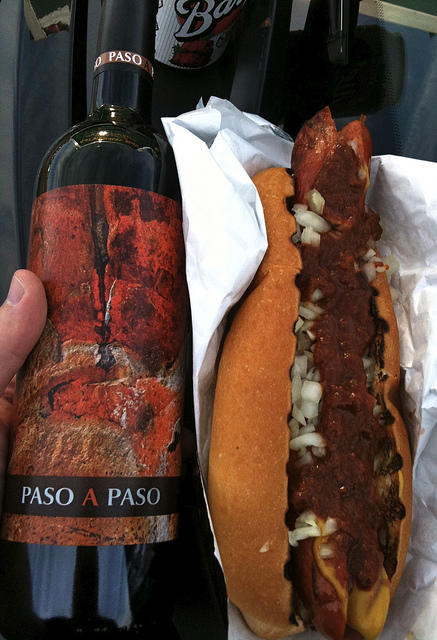}
    \includegraphics[width=0.9\textwidth,height=0.8\textwidth]{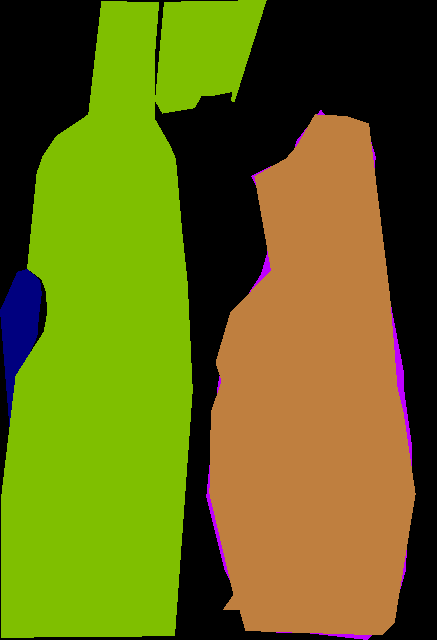}
    \includegraphics[width=0.9\textwidth,height=0.8\textwidth]{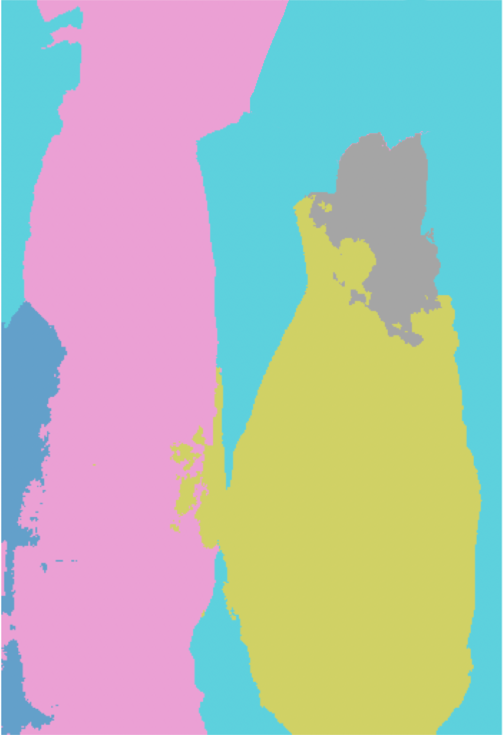}
    \includegraphics[width=0.9\textwidth,height=0.8\textwidth]{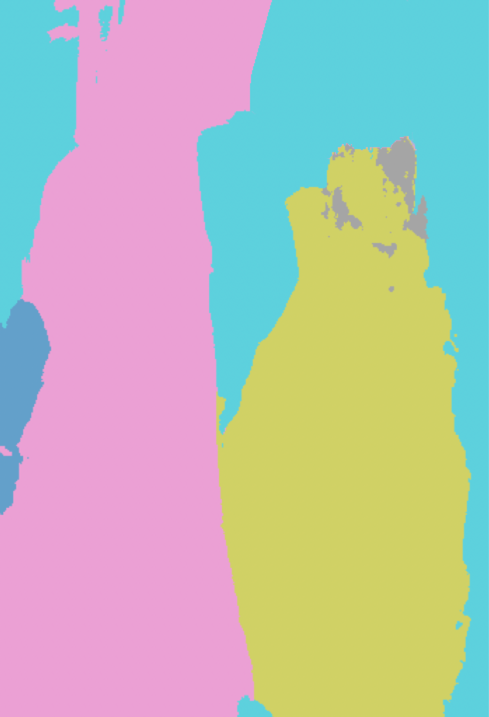}
    \label{fig:short-6e}
        \end{minipage}}
\caption{Qualitative segmentation results on the validation set of MS COCO 2014. From top to down: original image, ground-truth, SIPE \cite{Chen02909} and our eX-ViT.}
\label{fig:fig6}
\end{figure}

\begin{table}
  \caption{Performance comparison of state-of-the-art WSSS methods in terms of mIoU (\%) on the MS COCO 2014 validation set. $Sup.$ indicates supervision type. $\mathcal{I}$: image-level labels; $\mathcal{S}$: saliency maps.}
  \centering
  \setlength{\tabcolsep}{2pt}
  \begin{tabular}{l|l|l|c}
  \toprule  % 顶部线 
  Method & \makebox[0.1\textwidth][l]{$Sup.$} & \makebox[0.1\textwidth][l]{Backbone} & \makebox[0.1\textwidth][c]{mIoU (\%)} \\
  \midrule  % 中部线
  (CVPR'18) DSRG \cite{HuangWWLW18} & $\mathcal{I} + \mathcal{S}$ & VGG16 & 26.0 \\
  (ICCV'21) AuxSegNet \cite{XuOBBS021} & $\mathcal{I} + \mathcal{S}$ & ResNet38 & 33.9 \\
  (CVPR'21) EPS \cite{LeeLLS21} & $\mathcal{I} + \mathcal{S}$ & ResNet101 & 35.7 \\
  (NeurIPS'20) CONTA \cite{ZhangZT0S20} & $\mathcal{I}$ & ResNet101 & 33.4 \\
  (CVPR'20) SEAM \cite{WangZKSC20} & $\mathcal{I}$ & ResNet38 & 31.9 \\
  (ICCV'21) CSE \cite{KweonYKPY21} & $\mathcal{I}$ & ResNet38 & 36.4 \\
  (ICCV'21) CDA \cite{SuSLW21} & $\mathcal{I}$ & ResNet38 & 33.2 \\
  (CVPR'22) SIPE \cite{Chen02909} & $\mathcal{I}$ & ResNet101 & 40.6 \\
  (CVPR'22) ReCAM \cite{chencrea962} & $\mathcal{I}$ & ResNet101 & 39.4 \\
  (CVPR'22) Ru et al. \cite{ruaffinity2022}  & $\mathcal{I}$ & MiT-B1 & 38.9 \\
  \bf{(Ours) eX-ViT} & $\mathcal{I}$ & ResNet38 & \bf{42.9} \\
  \bottomrule
  \end{tabular}
\label{table:tab3}
\end{table}

\subsubsection{Comparison on segmentation results}
The comparison results among the fully-supervised and weakly supervised state-of-the-art methods on PASCAL VOC 2012 validation and test sets are reported in \cref{table:tab2}. We use the generated pseudo semantic segmentation labels to train DeepLabV2 \cite{ChenPKMY18}. Among the compared methods, the eX-ViT is able to remarkably improve the segmentation performance using only image-level labels on the validation and test sets, respectively. As can be observed, compared to the fully-supervised methods, the eX-ViT is able to obtain comparable performance with 71.2\% mIoU on validation set and 71.1\% mIoU on test set. Compared with the recent state-of-the-art weakly supervised models, e.g., EPS \cite{LeeLLS21} and EDAM \cite{WuHGWWLL21} which use both additional saliency maps and image-level labels as supervision signals, eX-ViT still shows superior performance. The qualitative segmentation results on the validation set are shown in \cref{fig:fig5}. Based on our model, DeepLabV2 can produce accurate and complete object segmentation results in various challenging scenarios, including different object scales and multiple objects.

In \cref{table:tab3} we report the semantic segmentation results on the MS COCO 2014 dataset. It is observed that methods with the supervision of saliency maps such as DSRG \cite{HuangWWLW18} and AuxSegNet \cite{XuOBBS021} do not provide results comparable or superior to the WSSS methods with only image-level labels. The poor performance is caused by the limitation of saliency maps generated by pre-trained models. Instead, our method that leverages image-level labels achieves a segmentation mIoU of 42.9\% with ResNet38 backbone, which surpasses most recent state-of-the-art WSSS methods including SEAM \cite{WangZKSC20}, CSE \cite{KweonYKPY21}, and CDA \cite{SuSLW21} by a large margin. Several qualitative segmentation results are shown in \cref{fig:fig6}. These results confirm the effectiveness of our model, which is consistent with our intuition. Specifically, our eX-ViT remarkably improves the overall performance with the indispensable block of E-MHA and the AttE module. Adding these modules explicitly encourages eX-ViT to gain insightful clues on the complete object scene, and boost the model efficiency on producing accurate and complete object boundaries. 

\subsubsection{Comparison on interpretability}
To compare our method with other explainable methods, we also adopt another two common metrics, i.e., average precision (AP) and average recall (AR). Which are commonly used in the literature to measure the interpretability \cite{KhoslaRTO15}. We evaluate our method using the Deit backbone \cite{GuillauminKF14} and conduct the weakly-supervised image segmentation experiments, which is in line with earlier work \cite{CheferGW21}. The quantitative results are shown in \cref{table:tab4}. We can see that our model clearly surpasses the ViT model which contains the raw attentions, it reveals that our MAXNet achieves an AP of 15.7\%, and an AR of 22.3\%. We also observe that the post-hoc interpretability methods such as Rollout \cite{AbnarZ20}, GradCAM \cite{selvaraju2017grad} and partial LRP \cite{BinderMLMS16} do not obtain faithful results compared to the counterparts. Which is possibly caused due to the extensive noises introduced by gradients or propagation rules.

\begin{figure}
  \centering
  \setlength{\abovecaptionskip}{0.cm}
  \setlength{\belowcaptionskip}{0.cm} 
  \subfloat[Image]{
        \begin{minipage}[b]{0.19\textwidth}
     \includegraphics[width=1\textwidth,height=0.8\textwidth]{figs/coco/COCO_val2014_000000168337.jpg}
     \includegraphics[width=1\textwidth,height=0.8\textwidth]{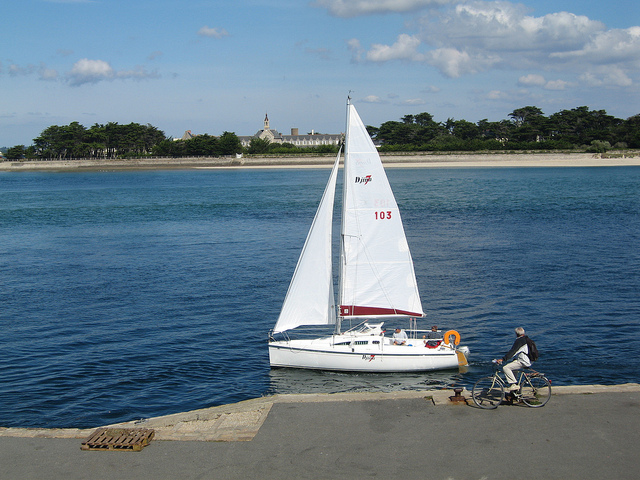}
     \includegraphics[width=1\textwidth,height=0.8\textwidth]{figs/coco/COCO_val2014_000000057645.jpg}\vspace{-12pt}
    \label{fig:short-5a}
        \end{minipage}}
        \hspace{-5pt}
    \subfloat[Ground-truth]{
        \begin{minipage}[b]{0.19\textwidth}
   \includegraphics[width=1\textwidth,height=0.8\textwidth]{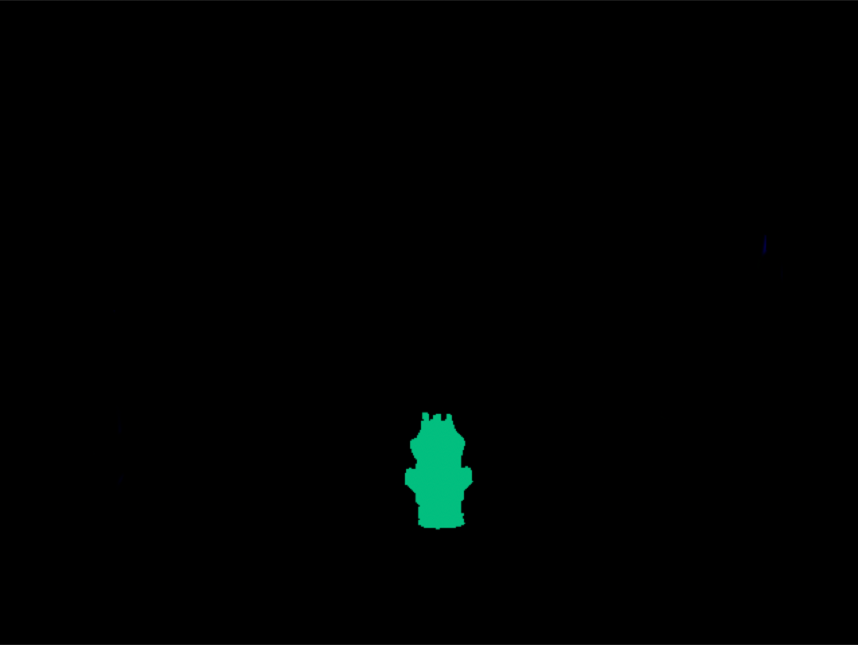}
    \includegraphics[width=1\textwidth,height=0.8\textwidth]{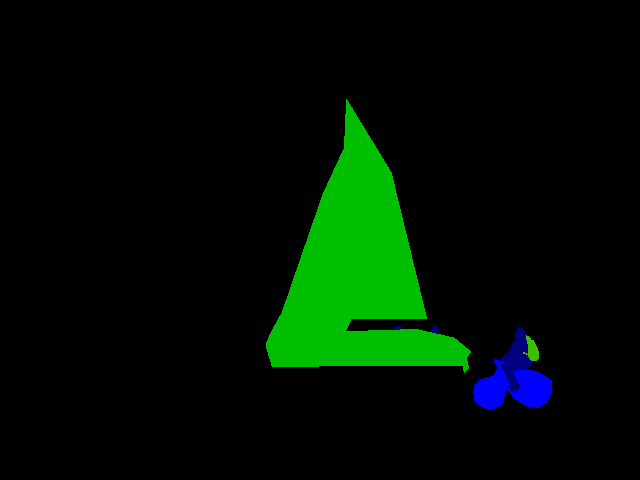}
    \includegraphics[width=1\textwidth,height=0.8\textwidth]{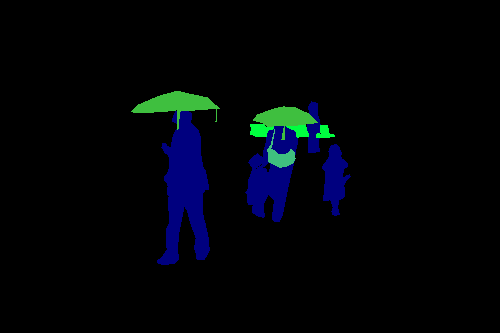}\vspace{-12pt}
    \label{fig:short-5b}
        \end{minipage}}
        \hspace{-5pt}
    \subfloat[E-MHA]{
        \begin{minipage}[b]{0.19\textwidth}
    \includegraphics[width=1\linewidth, height=0.8\linewidth]{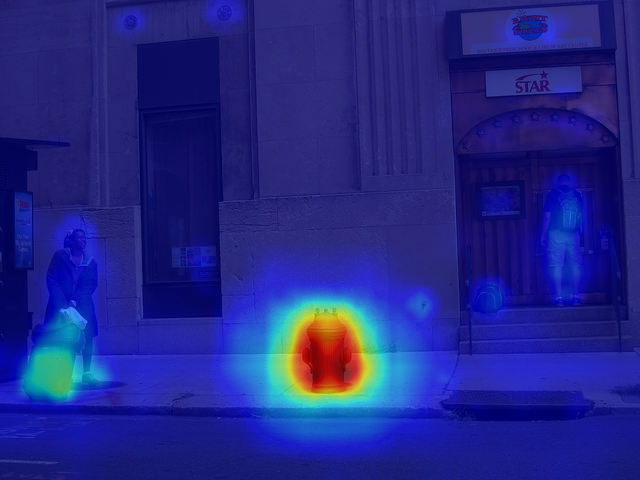}
    \includegraphics[width=1\textwidth,height=0.8\textwidth]{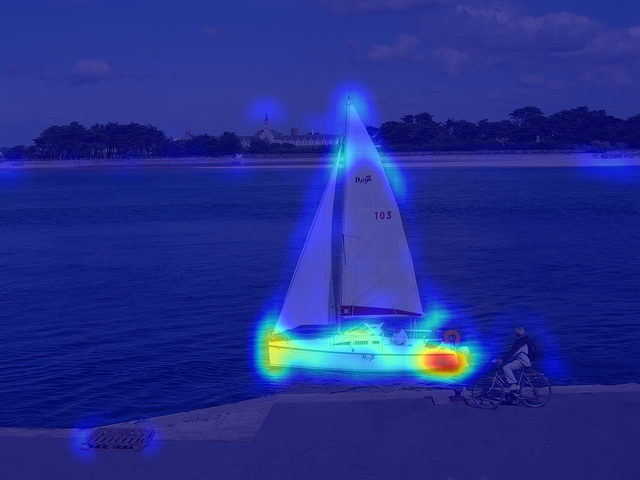}
    \includegraphics[width=1\textwidth,height=0.8\textwidth]{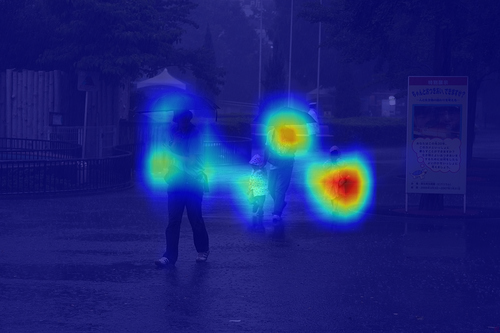}\vspace{-12pt}
    \label{fig:short-5c}
        \end{minipage}}
        \hspace{-5pt}
    \subfloat[w/o AttE]{
        \begin{minipage}[b]{0.19\textwidth}
     \includegraphics[width=1\linewidth, height=0.8\linewidth]{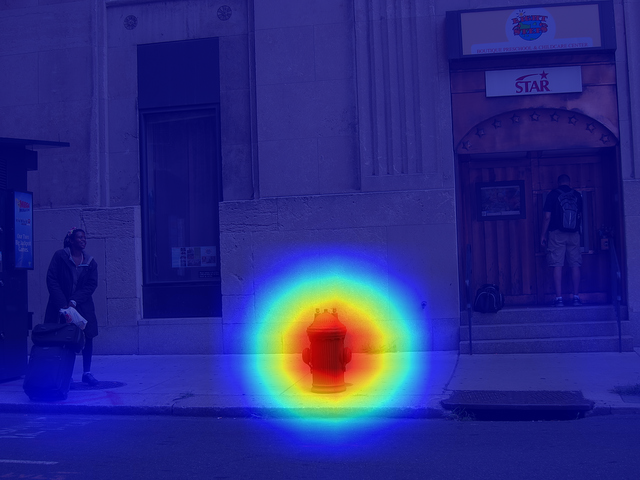}
     \includegraphics[width=1\textwidth,height=0.8\textwidth]{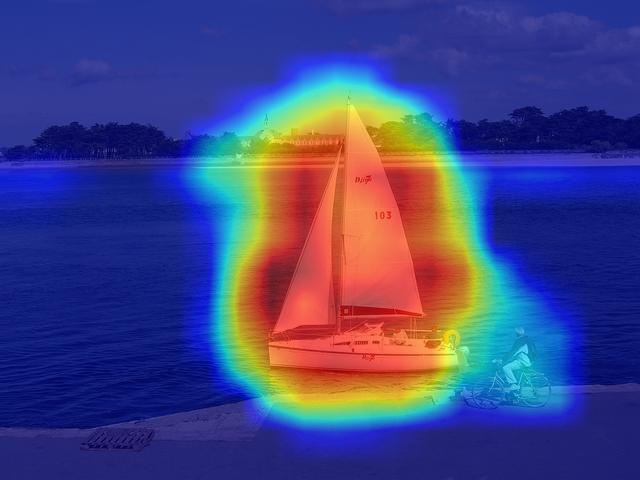}
    \includegraphics[width=1\textwidth,height=0.8\textwidth]{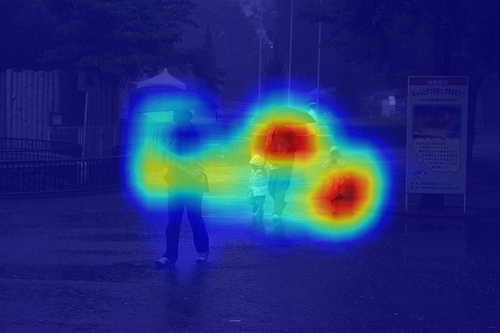}\vspace{-12pt}
    \label{fig:short-5c}
        \end{minipage}}
        \hspace{-5pt}
    \subfloat[w/ AttE]{
        \begin{minipage}[b]{0.19\textwidth}
     \includegraphics[width=1\linewidth, height=0.8\linewidth]{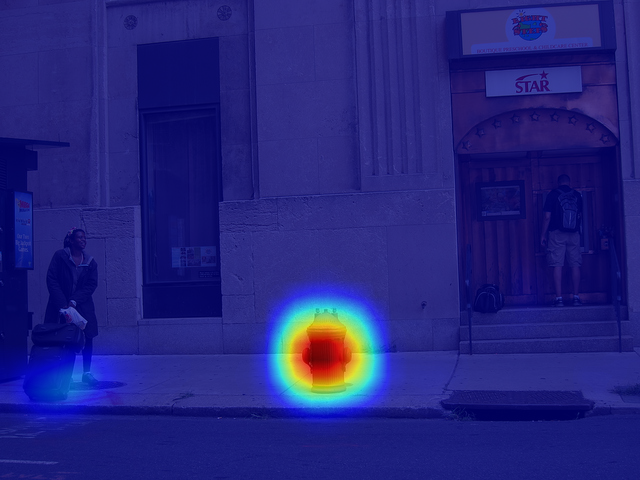}
     \includegraphics[width=1\textwidth,height=0.8\textwidth]{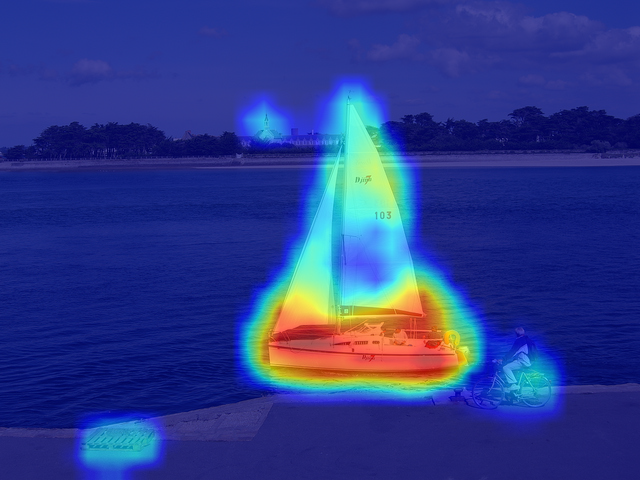}
    \includegraphics[width=1\textwidth,height=0.8\textwidth]{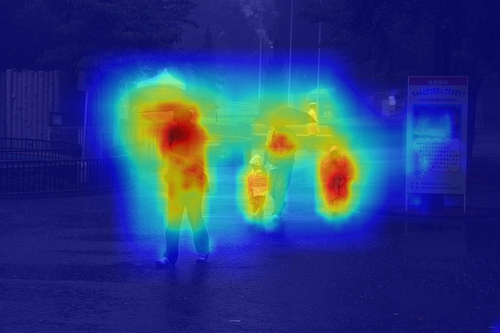}\vspace{-12pt}
    \label{fig:short-5c}
        \end{minipage}}
\caption{Visualization results on the MS COCO 2014 validation set.}
\label{fig:fig7}
\end{figure}

\cref{fig:fig7} shows three cases of visualization results along with their ground truth segmentation label maps. Compared to the original CAM without AttE, attention maps produced with our model perform well on precisely locating both small and large objects with more complete object boundaries. This verifies our intuition with the design of eX-ViT and suggests that our proposed model is effective on learning comprehensive features for complete target objects.

\begin{table}[!t]
  \caption{Performance comparison of various methods on the MS COCO validation set.}
  \centering
  \setlength{\tabcolsep}{2pt}
  \begin{tabular}{l|c|c|c|c|c|c}
  \toprule  % 顶部线 
  \makebox[0.2\textwidth][c]{Method} & \makebox[0.1\textwidth][c]{AP} & \makebox[0.1\textwidth][c]{AP\_{medium}} & \makebox[0.1\textwidth][c]{AP\_{large}} & \makebox[0.1\textwidth][c]{AR} & \makebox[0.1\textwidth][c]{AR\_{medium}} & \makebox[0.1\textwidth][c]{AR\_{large}} \\
  \midrule  % 中部线
  (ICCV'17) GradCAM \cite{selvaraju2017grad} & 2.3 & 2.3 & 4.7 & 5.5 & 5.9 & 10.7 \\
  (ACL'19) Partial LRP \cite{VoitaTMST19} & 4.7 & 8.0 & 5.1 & 10.4 & 19.9 & 8.0 \\
  (ICLR'20) ViT \cite{dosovitskiy2020ViT} & 5.6 & 9.6 & 6.9 & 11.7 & 21.8 & 10.8 \\
  (ACL'20) Rollout \cite{AbnarZ20}  & 0.1 & 0.1 & 0.2 & 0.4 & 0.1 & 0.9 \\
  (CVPR'21) Trans. attribution \cite{CheferLRP21} & 7.2 & 10.4 & 12.4 & 13.4 & 21.0 & 19.4 \\
  (ICCV'21) Chefer et al. \cite{CheferGW21} & 13.1 & 14.4 & 24.6 & 19.3 & 23.9 & 33.2 \\
  \bf{(Ours) eX-ViT} & \bf{15.7} & \bf{15.3} & \bf{26.5} & \bf{22.3} & \bf{24.3} & \bf{36.1} \\
  \bottomrule
  \end{tabular}
\label{table:tab4}
\end{table}

\subsection{Ablation studies}
This section presents ablation studies to analyze the contributions of each component in our eX-ViT, including the transformer encoder with the proposed Explainable Multi-Head Attention (E-MHA), the Attribute-guided Explainer (AttE), the global-level attribute-guided loss function $\mathcal{L}_{\rm global}$, the local-level attribute discriminability loss function $\mathcal{L}_{\rm dis}$, and the attribute diversity loss function $\mathcal{L}_{\rm div}$.

\subsubsection{Effectiveness of E-MHA}
\label{sec4.3}
It is an intuition that improved transformer attention mechanism in E-MHA will improve model's ability to generate pseudo segmentation labels. In order to verify this idea, we simply apply CAM to the last K transformer encoder layers, denoted as E-MHA-CAM. \cref{table:tab5} reports the mIoU results of the pseudo labels generated by CAM with the backbone of ResNet38, ResNet50, Conformer-S \cite{PengHGXWJY21}, and the encoder $E^{\theta}$ in our proposed eX-ViT. As can be seen, the backbone with the E-MHA module shows superior performance to its CNN counterparts. Specifically, E-MHA-CAM achieves the mIoU of 52.31\%, which is a significant gain of +4.92\%, +4.01\% over ResNet38-CAM and ResNet50-CAM, respectively. By comparing the E-MHA with the recent state-of-the-art architecture, i.e., Conformer-S \cite{PengHGXWJY21}, we find that our proposed E-MHA still achieves a promising result. In details, compared to CrossFormer-S \cite{PengHGXWJY21} which explicitly uses multi-scale representations with convolutions to localize object details, E-MHA-CAM achieves the best mIoU of 52.31\%, which is 0.61\% points higher than CrossFormer-S-CAM. The performance improvement shows that exploiting the most frequent and robust features by use of E-MHA is highly effective for WSSS tasks which require discriminative features to localize instances.

\begin{table}
  \caption{Performance comparison of various methods in terms of mIoU (\%) on the PASCAL VOC 2012 training set.}
  \centering
  \setlength{\tabcolsep}{2pt}
  \begin{tabular}{l|c}
  \toprule  % 顶部线 
  \makebox[0.3\textwidth][c]{Method} & \makebox[0.1\textwidth][c]{mIoU(\%)} \\
  \midrule  % 中部线
  (CVPR'19) ResNet50-CAM \citet{AhnCK19} & 48.30 \\
  (CVPR'20) ResNet38-CAM \cite{WangZKSC20} & 47.43 \\
  (ICCV'21) Conformer-S-CAM \cite{PengHGXWJY21} & 51.70 \\
  \bf{(Ours) E-MHA-CAM} & \bf{52.31} \\
  \bottomrule
  \end{tabular}
\label{table:tab5}
\end{table}

\begin{table}[!t]
  \caption{Effect of the contributions from various modules in terms of mIoU (\%) on the PASCAL VOC training set.}
  \centering
  \setlength{\tabcolsep}{2pt}
  \begin{tabular}{cccc|c|c}
  \toprule  % 顶部线 
  \makebox[0.07\textwidth][c]{$E^{\theta}$} &
  \makebox[0.07\textwidth][c]{\makebox[0.1\textwidth][c]{$\mathcal{L}_{\rm global}$}} & \makebox[0.12\textwidth][c]{$\mathcal{L}_{\rm dis}$} & \makebox[0.07\textwidth][c]{$\mathcal{L}_{\rm div}$} & \makebox[0.1\textwidth][c]{training} & \makebox[0.1\textwidth][c]{validation} \\
  \midrule  % 中部线
  \checkmark &  & & & 44.72 & 50.20 \\
  \checkmark & \checkmark & & & 53.71 & 55.43 \\
  \checkmark &  & \checkmark & & 51.25 & 54.63 \\
  \checkmark &  &  & \checkmark & 52.13 & 55.50 \\
  \checkmark & \checkmark & \checkmark & & 55.27 & 58.10 \\
  \checkmark & \checkmark &  & \checkmark & 58.08 & 59.82 \\
  \checkmark & \checkmark & \checkmark & \checkmark & \bf{59.13} & \bf{61.20} \\
  \bottomrule
  \end{tabular}
\label{table:tab6}
\end{table}

\subsubsection{Effectiveness of the AttE and attribute-guided loss}
\cref{table:tab6} gives an ablation study of each component in our proposed eX-ViT. We consider the first row as a baseline, where the results of the object localization maps are obtained via the CAM approach. As is observed from the table, the baseline can be further improved to 53.71\% and 55.43\% on the training and validation set, respectively by using the attribute features obtained via AttE to refine the learned transformer attention from the eX-ViT. Empirically, with attribute-guided discriminability loss $\mathcal{L}_{\rm dis}$ the pseudo segmentation label maps can be improved by +6.53\% compared to the baseline on the PASCAL VOC training set (51.25\% vs. 44.72\%) even without the global supervision $\mathcal{L}_{\rm global}$. Moreover, the $\mathcal{L}_{\rm dis}$ further improves the mIoU to 54.63\% on the validation set. By incorporating the attribute diversity loss function $\mathcal{L}_{\rm div}$ to explicitly regularize the attribute structure of the feature space, our full model gains promising results. Particularly, the results in \cref{table:tab6} indicate that the proposed model performs better with the diversity constraint $\mathcal{L}_{\rm div}$ on the local consistency, which brings +4.37\% and 4.39\% mIoU improvements on the training and validation sets, respectively compared to the global-level loss. This also confirms our theory that improving the diversity among attributes promotes the learning of comprehensive localization maps.

\begin{figure}[t]
  \centering
  \includegraphics[width=0.8\linewidth, height=0.5\linewidth]{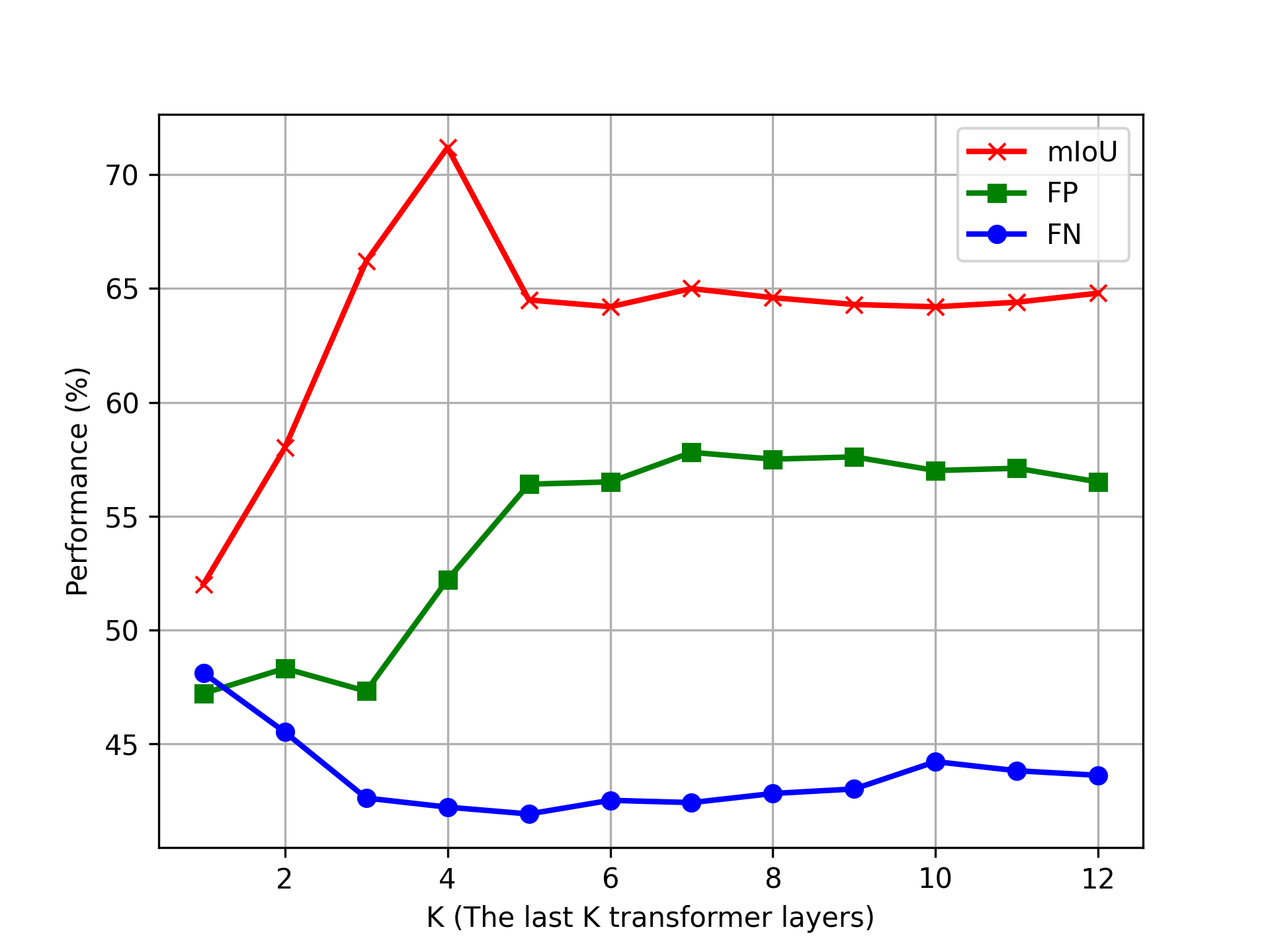}
  \caption{Evaluation of object localization maps generated by fusing the class-specific attentions from the last $K$ transformer layers in eX-ViT's encoder $E^{\theta}$ in terms of false positives (FP), false negatives (FN) and mIoU. The larger FP and FN values denote have more over-activated pixels, while the higher mIoU value indicates the generated localization maps have less over-activated pixels and more complete object coverage.}
  \label{fig:fig8}
\end{figure}

\subsubsection{Influence of the number of fused transformer layers}
We further explore the impact of the number of fused transformer layers in \cref{eq10} on the PASCAL VOC training set. Following the common practice in the prior work \cite{WangZKSC20}, we adopt three metrics to evaluate the performance, i.e., false positives (FP), false negatives (FN), and mIoU. The larger FP and FN values denote higher degrees of over-activated and under-activated areas, respectively. In \cref{fig:fig8}, we compare the performance of the model variants using different numbers of the fused transformer layers. As is observed, when fusing layers with more than 10, we obtain localization maps with a larger ${FN}$ value, which suggests the generated localization maps have more over-activated pixels and less complete activation coverage. This is mainly due to the limited ability of lower layers to encode high-level representations. By reducing the number of fused layers from the encoder $E^{\theta}$, the performance of predicted localization maps become much better, i.e., more complete activation coverage (lower ${FN}$ value) or fewer over-activated regions (lower ${FP}$ value). Overall, the evaluation results indicate that using the last three attention layers can achieve the best mIoU of 71.2\% with lower ${FN}$ and ${FP}$ values. Therefore we set $K=3$ throughout the paper.  

\subsubsection{Influence of the number of attributes}
\label{sec:numberc}
The attribute-guided scheme allows the model to encode richer semantics into each attribute feature at a granular level. In order to discover the most suitable $S$ concerning different datasets, we conduct extensive experiments to compare the performance of the model variants with different settings of hidden dimension $c$ and the number of attributes $S$. As shown in \cref{table:tab7}, when $c=128$, the model learns weaker representations for both datasets. In contrast, the performance becomes much better when the hidden dimension is enlarged to be 512. However, as the number of attributes increases to 16, the model exhibits a poor mIoU accuracy. In the end, we find that $c=256$ achieves consistently superior performance across a range of attribute numbers. The best performance is achieved when $S=8$ on the PASCAL VOC 2012 $val$ set, and $S=16$ on the MS COCO 2014 $val$ set. These observations suggest that images in MS COCO 2014 tend to contain more local features that are discriminative for object localization.

\begin{table}
  \caption{The influence of the number of attributes in terms of mIoU (\%) on the PASCAL VOC and MS COCO 2014 validation sets.}
  \centering
  \setlength{\tabcolsep}{2pt}
  \begin{tabular}{c|c|c|c}
  \toprule  % 顶部线 
  \makebox[0.1\textwidth][c]{$c$} & \makebox[0.1\textwidth][c]{$S$} & \makebox[0.25\textwidth][c]{PASCAL VOC 2012 $val$} & \makebox[0.2\textwidth][c]{MS COCO 2014 $val$}\\
  \midrule  % 中部线
  512 & 8 & 69.42 & 40.31 \\
  512 & 16 & 69.03 & 38.92 \\
  \midrule
  256 & 4 & 63.48 & 36.69 \\
  256 & 8 & \bf{71.23} & 41.23 \\
  256 & 16 & 70.29 & \bf{42.92} \\
  \midrule
  128 & 4 & 68.63 & 37.25 \\
  128 & 8 & 68.56 & 38.79 \\
  \bottomrule
  \end{tabular}
\label{table:tab7}
\end{table}

\section{Conclusion}
\label{sec5}
In this paper, we proposed the eX-ViT, a new explainable vision transformer for weakly supervised semantic segmentation. In our framework, a novel Explainable Multi-Head Attention (E-MHA) module is proposed to produce discriminative feature representations with inherent explainability and noise robustness. Which is achieved by optimizing the dynamic alignment between the input tokens and attention weights. Moreover, a new Attribute-guided Explainer (AttE) module is designed to decompose the attention maps into the contribution of each individual attribute, empowering the feature representation with a set of attribute maps at a granular level. Based on AttE, we develop a self-supervised attribute-based loss to guide the learning of attribute features with the attribute discriminability mechanism and attribute diversity mechanism, which promotes the generation of diverse and discriminative object attributes. Extensive experiments were presented to demonstrate that the eX-ViT surpasses the state-of-the-art CNNs and transformers on two well-known benchmarks. We hope that the eX-ViT's superior performance on WSSS tasks will inspire future research on the exploitation of the explainability of transformers. 

\bibliography{Neural_Networks_YuLu}

\end{document}